\documentclass[runningheads]{llncs}
\usepackage{graphicx}
\usepackage{amsmath,amssymb} % define this before the line numbering.
\usepackage{color}
\usepackage[dvipsnames]{xcolor}
\usepackage[scientific-notation=true]{siunitx}

\usepackage{amsmath,amssymb} % define this before the line numbering.
\usepackage{dsfont}
\usepackage{pifont}
\usepackage{booktabs}
\usepackage{multirow}
\usepackage{xspace}
\usepackage[pagebackref=true,breaklinks=true,colorlinks,bookmarks=false]{hyperref}
\usepackage{cleveref}
\usepackage{wrapfig}

\makeatletter
\DeclareRobustCommand\onedot{\futurelet\@let@token\@onedot}
\def\@onedot{\ifx\@let@token.\else.\null\fi\xspace}

\makeatother

\newcommand{\cmark}{\ding{51}}%
\newcommand{\xmark}{\ding{55}}%

\newcommand{\ie}{i.e.}
\newcommand{\eg}{e.g.}

\newcommand{\transpose}[1]{#1^\top}

\begin{document}
\pagestyle{headings}
\mainmatter

% \def\ACCV22SubNumber{384}  % Insert your submission number here

%===========================================================
\title{Is an Object-Centric Video Representation Beneficial for Transfer? } % Replace with your title
% \titlerunning{ACCV-22 submission ID \ACCV22SubNumber}
% \authorrunning{ACCV-22 submission ID \ACCV22SubNumber}

% \author{Anonymous ACCV 2022 submission}
% \institute{Paper ID \ACCV22SubNumber}

\author{Chuhan Zhang\inst{1} \and
Ankush Gupta\inst{2} \and
Andrew Zisserman\inst{1}}

\authorrunning{C.~Zhang et al.}
\institute{Visual Geometry Group, Department of Engineering Science\\ University of Oxford\\
\email{\{czhang,az\}@robots.ox.ac.uk}\\
\and
DeepMind, London\\
\email{ankushgupta@google.com}}

% \maketitle
{\def\addcontentsline#1#2#3{}\maketitle}

%===========================================================
\begin{abstract}
    The objective of this work is to learn an {\em object-centric} video representation, with 
the aim of improving transferability to novel tasks, \ie, tasks different
from the pre-training task of action classification.
To this end, we introduce a new object-centric
video recognition model based on a transformer architecture. 
The model learns a set of object-centric  summary vectors for the video, and
uses these vectors to fuse the visual and spatio-temporal trajectory 
`modalities' of the video clip.  We also introduce a novel trajectory contrast
loss to further enhance objectness in these summary vectors. 

With experiments on four datasets---SomethingSomething-V2, SomethingElse, Action 
Genome and EpicKitchens---we show that the object-centric model outperforms prior
video representations (both object-agnostic and object-aware), when:
(1) classifying actions on unseen objects and unseen environments;  (2) low-shot learning of novel classes; (3) linear probe to other downstream tasks;  as well as 
(4) for standard action classification. 

    % \keywords{video action recognition, object centric representations, transfer learning}
\end{abstract}
%===========================================================

\section{Introduction}\label{sec:intro}

Visual data is complicated---a seemingly infinite
stream of events emerges from the interactions of a finite number of constituent {\em objects}. 
% Reality is a continuous explosion of interacting entities---a seemingly infinite
% stream of events emerges from a finite number of constituent objects. 
Abstraction and reasoning in terms of these entities and their 
inter-relationships---\emph{object-centric reasoning}---has long been argued by 
developmental psychologists to be a \emph{core} building block of infant 
cognition~\cite{spelke1992core}, and key for human-level common 
sense~\cite{tenenbaum2011grow}. 
% Human brain actively attends to object features for visual scene analysis~\cite{kanwisher2000visual}. 
This object-centric understanding posits
that objects exist~\cite{grill2005visual}, have permanence over time, and carry %persistent %% mass and shape can change...
along physical properties such as mass and shape that govern their interactions 
with each other. Factorizing the environment in terms of these objects as  recurrent 
entities  allows for combinatorial generalization in novel 
settings~\cite{tenenbaum2011grow}. Consequently, there has been a 
gradual growth in video models that embed object-centric inductive biases, \eg, 
augmenting the visual stream with actor or object bounding-box 
trajectories~\cite{orvit,stlt,sun2018actor,zhang2019structured}, graph algorithms 
on object nodes~\cite{strg,chen2019graph}, or novel architectures for efficient
discovery, planning and interaction~\cite{locatello2020object,battaglia2016interaction,kulkarni2019unsupervised}.

\begin{figure}[t]
  \centering
  \includegraphics[width=\linewidth]{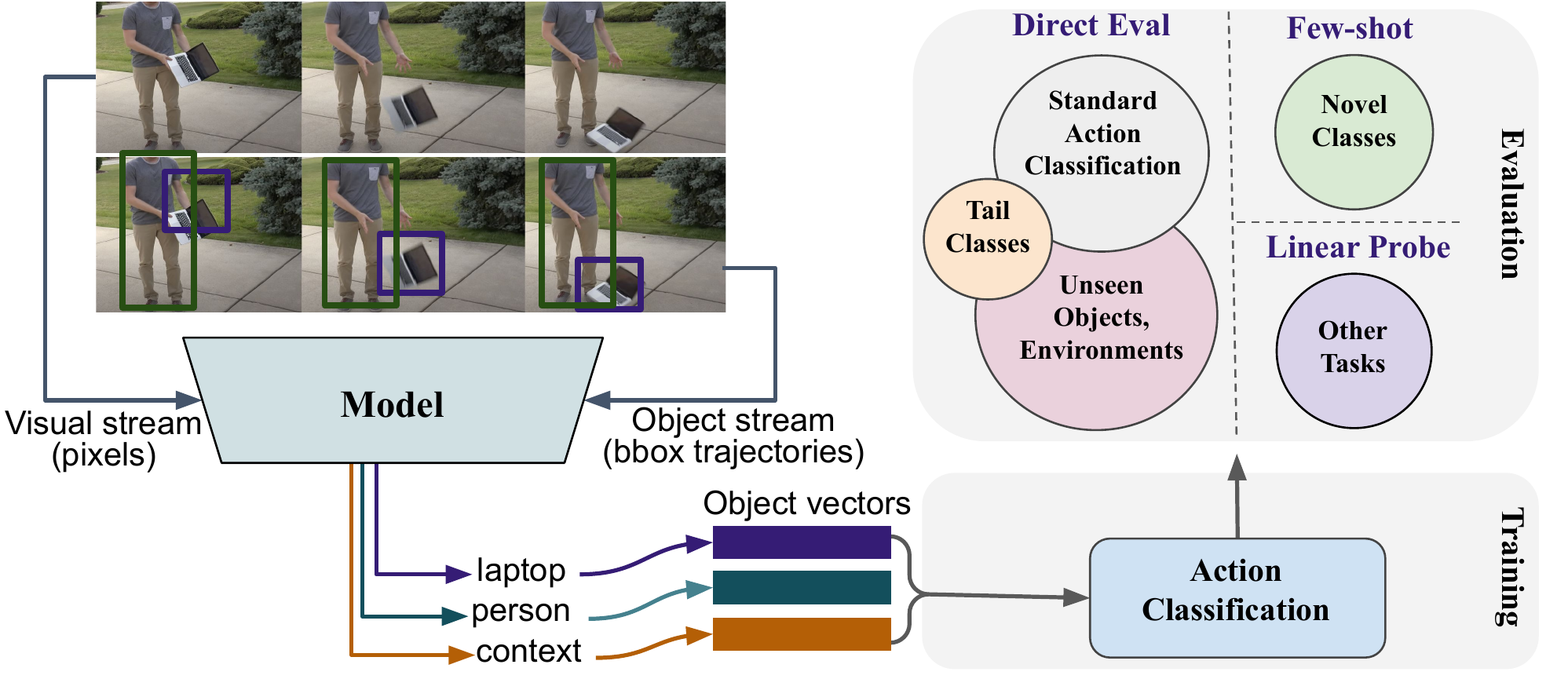}
  \caption{\textbf{Do Object-Centric video representations transfer better?}
  To induce objectness into visual representations of videos,
  we learn a set of 
  object-centric vectors---which are tied to specific objects present in the 
  video, as well as context vectors---which reason about relations and context.
The representation is built by
fusing the two 
  modality streams of the video---the visual stream,  and the spatio-temporal
  object bounding-box stream. 
  We train the model on the standard action recognition task, 
  and show that using the object and context vectors can lead to SOTA results 
  when evaluated for transfer to unseen objects, unseen environments, novel classes, other tasks and also 
  standard action classification.}
  \label{fig:teaser}
\end{figure}

The promise of object-centric representations is transfer across {\em tasks}.
Due to the shared underlying physics across different settings,
knowledge of object properties like shape, texture, and position can be 
repurposed with little or no modification for new 
settings~\cite{pathakICML18human}, much like infants who learn to manipulate objects and 
understand their properties, and then apply these skills to new objects or new 
tasks~\cite{gopnik2000scientist,smith2018developing}.

In this paper we investigate this promise by developing an {\em object-centric} video model and determining
if it has superior task generalization performance compared to 
object-agnostic and other recent object-centric methods. In a similar manner to pre-training a classification model
on ImageNet, and then using the backbone network for other tasks, we pre-train our object-centric model on 
the action recognition classification task, and then determine its performance on downstream tasks using a linear
probe.

We consider a model to be object-centric if it learns a set of object summary vectors,
that  \emph{explicitly} distil information about a specific object  into a particular latent variable. 
In contrast, in object-agnostic~\cite{carreira2017quo,lin2019tsm,xie2018rethinking,yang2020temporal} 
or previous state-of-the-art object-centric video models~\cite{orvit,stlt,sthelse}, the 
object information is de-localized throughout the representation.

To this end, we introduce a novel architecture based on  a transformer~\cite{vaswani2017attention}
that achieves an object-centric representation through its
design and its training.
First, a bottleneck representation is learned, where a set of object query 
vectors~\cite{carion2020end} tied to specific constituent objects, cross-attend in the manner of DETR~\cite{detr} to visual features and corresponding bounding-box trajectories.
We demonstrate this cross-attention based fusion is an effective 
method for merging the two modality streams~\cite{simonyan2014two,orvit,stlt}---visual
and geometric---complementing the individual streams.
We call this `modality' fusion module an \emph{Object Learner}.
Second, a novel \emph{trajectory contrast} loss is introduced to further enhance
object-awareness in the object summaries. Once learnt, this
explicit set of object summary vectors are repurposed and refined for downstream 
tasks.

% ** need to spell out more fully the ideas of transfer to different domains and 
% tasks (i.e. differentiate domains/tasks), with a few examples, since
% the task transfer will now be our main objective **

% transfer to: (1) smelse unseen objects, (2) few-shot, (3) hand-contact state 
%  action-genome human-object predicate prediction, (4) epic hand/object traj prediction.
We evaluate the task generalization ability of the object-centric video
representation using a number of challenging  transfer tasks and settings:
\begin{enumerate}
  \item \textbf{Unseen data:} Action classification with known actions (verbs),
  but novel objects (nouns) in the SomethingElse dataset~\cite{sthelse};
  Action classification with known actions (verbs and nouns), but unseen kitchens in EpicKitchens~\cite{epickitchen}.
  \item \textbf{Low-shot data:} Few-shot action classification in SomethingElse; Tail-class classification in EpicKitchens.
  \item \textbf{Other downstream tasks:} Hand contact state estimation in SomethingElse, and human-object predicate prediction in ActionGenome~\cite{ji2020action}. 
\end{enumerate}
Note, task 3 uses a linear probe on pre-trained representations for rigorously
quantifying the transferability.
In addition to evaluating the transferability as above, we also benchmark the
learned object-centric representations on the standard task of action classification.
\noindent In summary, our key contributions are:
\begin{enumerate}
  \item A new object-centric video recognition model with explicit object representations. 
  The object-centric representations are learned by using a novel cross-attention based module which fuses the visual and geometric 
  streams, complementing the two individually.
  \item The object-centric model sets a new record on a comprehensive set of tasks which evaluate transfer efficiency and accuracy on unseen objects, novel classes and new tasks on:  SomethingElse, Action Genome and EpicKitchens.
  \item Significant gains over the previous best results on 
  standard action recognition benchmarks: 74.0\%(+6.1\%) on SomethingSomething-V2, 66.6\%(+6.3\%) on Action 
Genome, and 46.3\%(+0.6\%) top-1 accuracy on EpicKitchens.
\end{enumerate}

\section{Related Work}\label{s:rel_work}

\noindent\textbf{Object-centric video models.}
Merging spatio-temporal object-level information and visual appearance for video
recognition models has been explored extensively.
These methods either focus solely on the human actors in the videos~\cite{sun2018actor,girdhar2019video,zhang2019structured}, or more generally model human-object interactions~\cite{gao2020drg,kato2018compositional,xu2019learning,gkioxari2018detecting}.
The dominant approach involves RoI-pooling~\cite{ren2015faster,he2017mask}
features extracted from a visual backbone using object/human bounding-boxes
generated either from object detectors~\cite{gupta2007objects}, or 
more generally using a region proposal network (RPN)~\cite{sun2018actor,baradel2018object,girdhar2019video,arnab2021unified,strg,sfi} on each frame independently, 
followed by global aggregation using recurrent models~\cite{ma2018attend}.
% For example, Wang~\etal~(STRG)~\cite{strg} construct a graph convolutional 
% network with object nodes obtained using RPN boxes, with similarity and 
% spatio-temporal relation modelling.
The input to these methods is assumed to just be RGB pixels, and the object
boxes are obtained downstream. A set of object-centric
video models~\cite{orvit,sthelse,stlt} assume object boxes as \emph{input}, and
focus on efficient fusion of the two streams; we follow this setting.
Specifically, ORViT~\cite{orvit} is an object-aware vision transformer~\cite{vit}
which incorporates object information in two ways: (1) by attending to 
RoI-pooled object features, and (2) by merging encoded trajectories
at several intermediate layers.
STIN~\cite{sthelse} encodes the object boxes and identity independently of 
the visual stream, and merges the two through concatenation before feeding 
into a classifier.
STLT~\cite{stlt} uses a transformer encoder on object boxes, first across all
objects in a given frame, and then across frames, before fusing with appearance
features. We adopt STLT's hierarchical trajectory encoder, and develop a
more performant cross-attention based fusion method.\\

\noindent\textbf{Multi-modal fusion.}
Neural network architectures which fuse multiple modalities, both within the visual domain, \ie, images and videos~\cite{slowfast} with optical flow~\cite{simonyan2014two,feichtenhofer2016convolutional}, bounding-boxes~\cite{sun2018actor,baradel2018object,girdhar2019video,arnab2021unified,strg,sfi}, as well as across other modalities, \eg, sound~\cite{arandjelovic2018objects,owens2018audio} and language~\cite{aytar2017see,frome2013devise,weston2011wsabie,sun2019videobert}, have been developed. The dominant approach was introduced in the classic two-stream fusion method~\cite{simonyan2014two} which processes the visual and optical flow streams through independent encoders before summing the final softmax predictions. Alternative methods~\cite{feichtenhofer2016convolutional} explore fusing at intermediate layers with different operations, \eg, sum, max, concatenation, and attention-based non-local operation~\cite{wang2018non}.
We also process the visual and geometric streams independently, but fuse using 
a more recent cross-attention based transformer decoder~\cite{mformer} 
acting on object-queries~\cite{carion2020end}.
An alternative to learning a single embedding representing all the input modalities, 
is to learn modality encoders which all map into the same joint vector space~\cite{alayrac2020self,nagrani2018learnable,aytar2017see}; such embeddings are 
primarily employed for retrieval.\\

\noindent\textbf{Benchmarks with object annotations.}
Reasoning at the \emph{object} level lies at the heart of computer vision, 
where standard benchmarks for recognition~\cite{deng2009imagenet},
detection and segmentation~\cite{everingham2015pascal,lin2014microsoft}, and
tracking~\cite{kristan2021ninth,huang2019got,dave2020tao,dendorfer2021motchallenge}
are defined for categories of objects.
Traditionally, bounding-box tracking of single~\cite{kristan2021ninth,huang2019got} 
or multiple objects~\cite{dave2020tao,dendorfer2021motchallenge}, or more 
spatially-precise video object segmentation~\cite{perazzi2016benchmark,xu2018youtube,yang2019video,wang2021unidentified} 
were the dominant benchmarks for object-level reasoning in videos.
More recently, a number of benchmarks probe objects in videos in other ways, 
\eg, ActionGenome~\cite{ji2020action} augments the standard action recognition 
with human/object based scene-graphs, SomethingElse~\cite{sthelse} tests for 
transfer of action recognition on novel objects, CATER~\cite{girdhar2020cater} 
evaluates compositional reasoning over synthetic objects, and 
CLEVERER~\cite{yi2019clevrer} for object-based counterfactual reasoning.\\

\noindent\textbf{Object-oriented reasoning.}
There is a large body of work on building in and reasoning with object-level 
inductive biases across multiple domains and tasks.
Visual recognition is typically \emph{defined} at the object-level both in images~\cite{lin2014microsoft,krishna2017visual,krishna2018referring,johnson2016densecap} and 
videos~\cite{gupta2007objects,saenko2012mid,ji2020action,xu2020spatio}.
Learning relations, expressed as edges, between entities/particles, expressed
as nodes in a graph has been employed for amortizing inference in simulators
and modelling dynamics~\cite{relationnet,battaglia2016interaction}.
Such factorized dynamics models conditioned on structured object representations
have been employed for future prediction and forecasting~\cite{ye2019compositional,liang2019peeking,wu2020future}.
Object-conditional image and video decomposition~\cite{pmlr-v97-greff19a,henderson2020unsupervised,locatello2020object,yang2021self} such as Monet~\cite{burgess2019monet} and Genesis~\cite{engelcke2019genesis}
 and generation~\cite{johnson2018image,herzig2018mapping,park2019gaugan,singh2019finegan,yang2021objectnerf,herzig2020learning}
methods benefit from compositional generalization.
Finally, object-level world-models have been used to constrain action-spaces,
and states in robotics~\cite{pmlr-v100-ye20a,devin2018deep} and control domains~\cite{pmlr-v97-bapst19a,kulkarni2019unsupervised,anand2019unsupervised}.

\section{An Object-Centric Video Action Transformer}\label{sec:model} 
We first describe the architecture of the object-centric video action recognition
model for fusing visual and trajectory streams. We then describe the training objectives for 
action classification and for learning the object representations. Finally, we
discuss our design choices, and the difference between our model and previous fusion methods,
and explain its advantages.

\begin{figure*}[t]
  \centering
  \includegraphics[width=\linewidth]{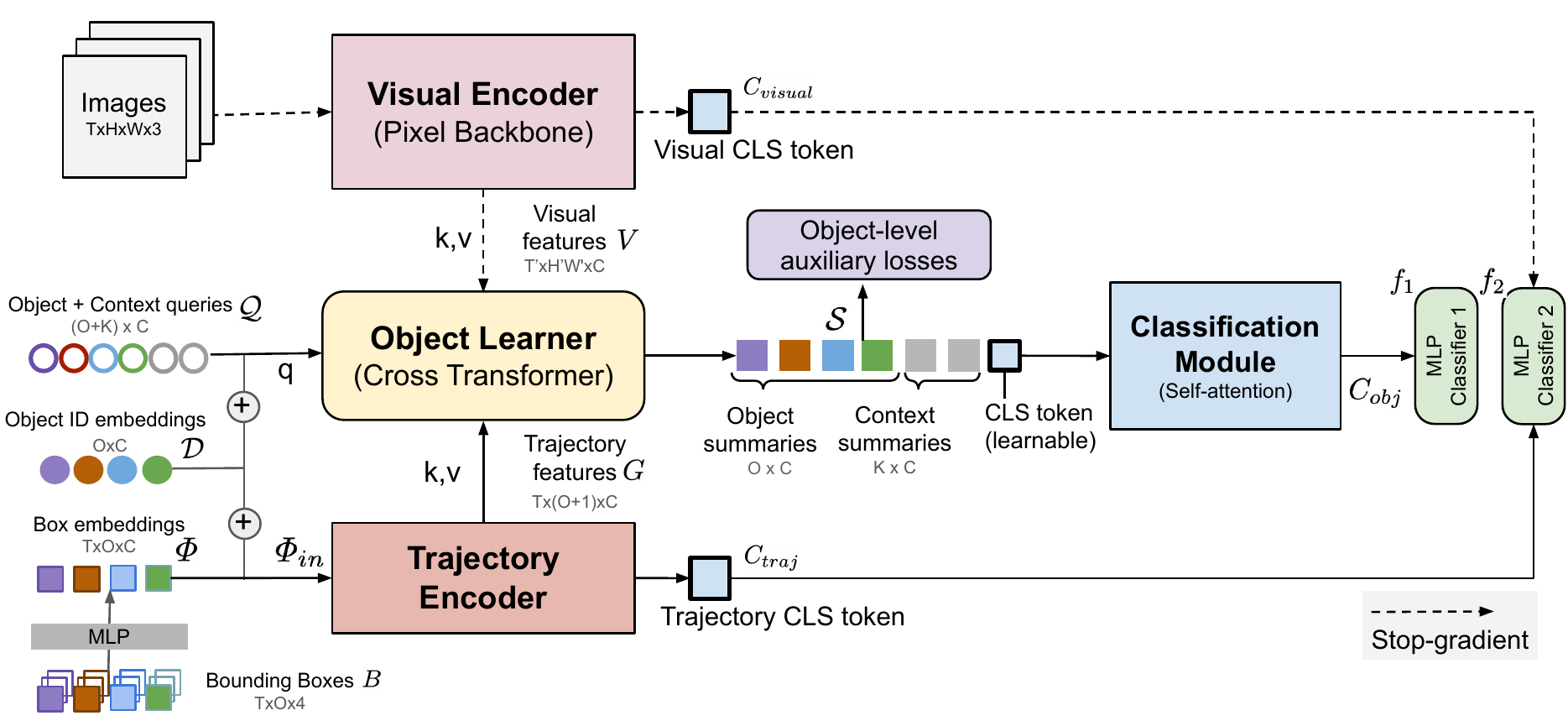}
  \caption{\textbf{The object-centric video transformer model architecture.} 
The {\em Visual Encoder} module ingests the RGB video and produces a set of object agnostic spatio-temporal tokens. 
The {\em Trajectory Encoder} module ingests the bounding boxes, and labels them with object ID embedddings $\mathcal{D}$, to produce object-aware trajectory spatio-temporal tokens. 
The {\em Object Learner} module fuses the visual and trajectory streams by querying with the object IDs $\mathcal{D}$, and outputs object summaries which contains both visual and trajectory information.
An object-level auxiliary loss is used to encourage each object summary vector to be tied to the object in the query. Finally, the {\em Classification} module ingests the outputs from the Object Learner to 
predict the class.
The model is trained with
cross-entropy losses applied to class predictions from the dual encoders
and Object Learner, together with an auxiliary loss.}
  \label{fig:arch-overview}
\end{figure*}

\subsection{Architecture}\label{s:arch}
The model is illustrated in \Cref{fig:arch-overview}, and consists of four transformer-based modules.
We briefly describe each module, with implementation details in section~\ref{s:impl}.\\

\noindent\textbf{Video Encoder.}
The encoder ingests a video clip $F$ of RGB frames ${F} = (f_1, f_2, 
\hdots, f_t)$, where $f_i \in \mathbb{R}^{ H{\times}W{\times}3}$.
The clip ${F}$ is encoded by a Video Transformer~\cite{mformer}
which tokenizes the frames by 3D patches to produce downsampled feature maps.
These feature maps appended with a learnable \texttt{CLS} token are processed by self-attention layers to obtain 
the spatio-temporal visual representations $V \in \mathbb{R}^{T'\times H'W'\times C}$ and video-level visual embedding $C_{visual}$.
We take the representation $V$ from the 6th self-attention layer to be the visual input of the Object Learner, and 
$C_{visual}$ from the last layer to compute the final loss in \cref{eq:loss}.\
% \agnote{(1)~We should state how $C_{visual}$ is used as is done for $C_{traj}$ below.
% (2)~We should also probably include something about how this encoder is used later on:
% The visual features $V$ and trajectory features (below) are input as keys and values to the Object Learner.}
% \agnote{If using intermediate features, state that here as well.}

\noindent\textbf{Trajectory Encoder.}
The encoder ingests the bounding box coordinates ${B^t} = (b^t_1, b^t_2, 
\hdots,b^t_o)$ of $\mathcal{O}$ number of annotated objects in the $t^\text{th}$ 
frame, where each box $b^t_i$ is in format $[x_1,y_1,x_2,y_2]$. These boxes are encoded into corresponding
box embeddings ${\Phi^t} = (\Phi^t_1, \Phi^t_2, \hdots,\Phi^t_o)$ through an MLP.
Object ID embeddings $\mathcal{D}=\{d_i\}_{i=1}^O$ are added to ${\Phi}$ to produce the sum ${\Phi_{in}}$, This is to keep the ID information persistent throughout the video length T; 
${\Phi_{in}}$ serves as the input to the Trajectory Encoder, which is a Spatial Temporal 
Layout Transformer (STLT) of~\cite{stlt}. STLT consists of two self-attention Transformers in sequence -- a Spatial Transformer and a Temporal Transformer.
First, the Spatial Transformer encodes boxes in every frame separately. 
It takes a learnable \texttt{CLS} token and box embeddings ${\Phi_{in}^t} \in \mathbb{R}^{O \times C}$ from frame $t$ as the input into the self-attention layers,
and output a frame-level representation ${l^t} \in \mathbb{R}^{1\times C}$ and spatial-context-aware box embeddings ${\Phi_{out}^t} \in  \mathbb{R}^{O \times C}$ respectively.
The Temporal Transformer models trajectory information over frames, it applies self-attention on the frame-level embeddings ${L} = (l^1, l^2, l^3, \hdots,l^T)$ from the Spatial Transformer with another learnable \texttt{CLS} token. 
Its output are temporal-context-aware frame embeddings ${L_{out}} \in \mathbb{R}^{T \times C}$  and a video-level trajectory representation ${C_{traj}} \in \mathbb{R}^{1\times C}$. 
$C_{traj}$ is used to compute the final loss in \cref{eq:loss}, while ${L_{out}} \in \mathbb{R}^{T \times 1 \times C}$ is concatenated with the 
${\Phi_{out}} \in \mathbb{R}^{T \times O \times C}$ from the Spatial Transformer to be the spatio-temporal trajectory embeddings 
$G \in \mathbb{R}^{T\times (O+1)\times C}$. $G$ is then downsampaled temporally from $T$ to $T'$, to be consistent with the temporal dimensions of the visual features. 
The faetures from the two encoders are then concatenated and used as trajectory input to the Object Learner. (See Supp. for detailed architecture.)

\begin{figure}[t]
  \centering
  \includegraphics[width=0.8\linewidth]{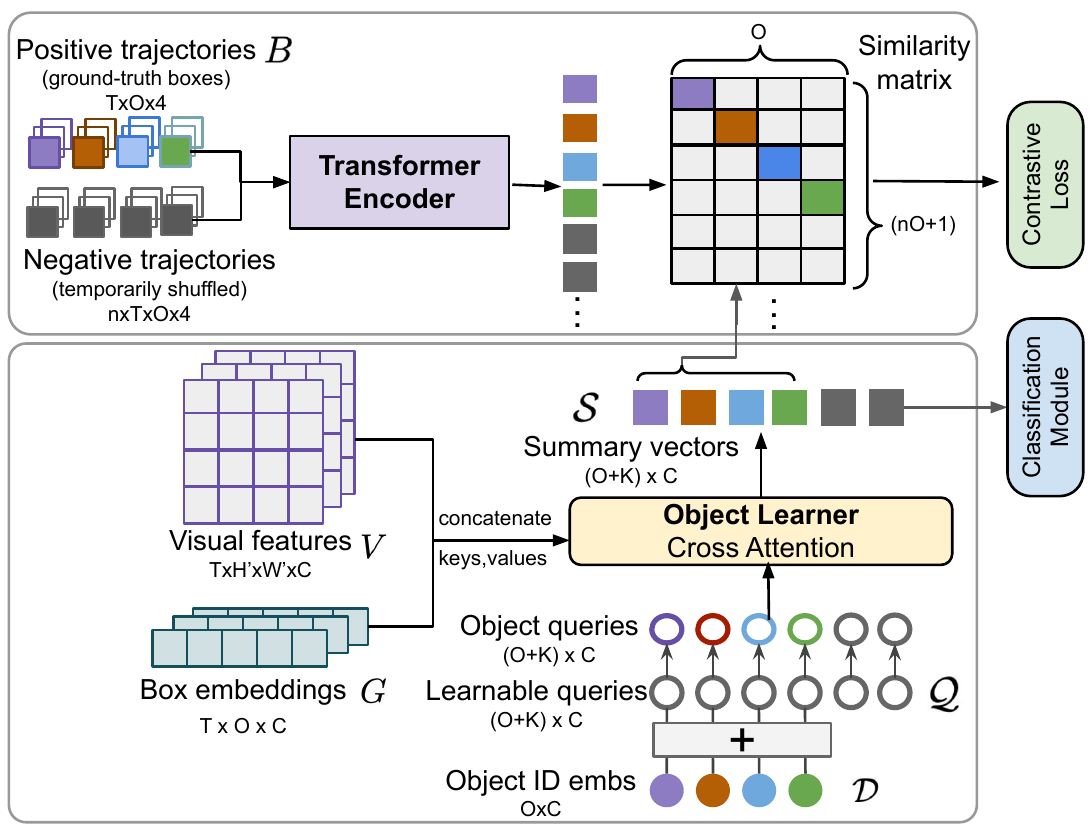}
  % \vspace{-3mm}
  \caption{\textbf{Object Learner and auxiliary loss.} The Object Learner is a cross-attention 
transformer, that outputs object-level summary vectors by attending to tokens from both the visual and 
trajectory encoders.
These summary vectors are used for downstream tasks like classification. An auxiliary loss is added where
the object  summary vectors are tasked with distinguishing GT and shuffled trajectory embeddings.}
  \label{fig:loss}
  % \vspace{-6mm}
\end{figure}

% $G \in \mathbb{R}^{T'\times O\times C}$ 
\noindent\textbf{Object Learner.}
The Object Learner module is a cross-attention
Transformer~\cite{vaswani2017attention} which has a query set
$\mathcal{Q}=\{q_i\}_{i=1}^{O+K}$ made up of $O$ learnable object
queries and $K$ learnable context queries. The same ID embeddings $\mathcal{D}$ from the Trajectory
Encoder are added to the first $O$ queries to provide object-specific
identification, while the remaining $K$ context queries can be learnt freely.
We concatenate the visual feature maps $V \in \mathbb{R}^{T'\times H'W'\times
C}$ and trajectory embeddings $G \in \mathbb{R}^{T'\times (O+1) \times C}$ as keys
and values in the cross-attention layers.
Note the query latents are video level (\ie, common across all frames),  and attend to the features from
the visual and trajectory encoders using cross-attention.  
% The cross-attention
% mechanism only reads from the two backbones without interfering the
% processing in them, thus keeping the two streams independent.
% Independent feature encoding in two modalities decouples visual and
% trajectory information, enables the model to obtain a better
% transferability between tasks and domains.  
The Object Learner outputs
summary vectors $\mathcal{S}=\{s_i\}_{i=1}^{O+K}$, $O$ of which are object centric, and the remaining $K$ carry context information.  
The output is independent of the number of video frames, with the visual and trajectory information distilled into the summary vectors.
\Cref{fig:loss} presents a schematic of the module.

\noindent\textbf{Classification Module.}
This is a light-weight cross-attention transformer that ingests the summary vectors output from the 
the Object Learner, together with a learnable query vector $C_{obj}$.
The vector output of this module is used for a linear classifier for the actionss
prediction.

% \vspace{-4mm}

\subsection{Objectives}\label{s:objectives}
We apply two types of losses to train the model. One is an object-level
auxiliary loss on the object summary vectors ${S} = (s_1, s_2,
\hdots,s_o)$ to ensure object-centric information is learned in these
vectors.  The other is a standard cross-entropy loss on action category prediction.\\

\noindent\textbf{Object-level Trajectory Contrast loss.}
The aim of this loss is to encourage the object specific latent queries in the Object Learner to attend to
both the trajectory and visual tokens, and thereby fuse the information from the two input modalities.
The key idea is to ensure that the video-level object queries do not ignore the identity, position encodings, and trajectory tokens.
The loss is a contrastive loss which encourages discrimination between the 
correct trajectories and others that are randomly perturbed or from other 
video clips in the batch.
This is implemented using an InfoNCE~\cite{infonce} loss, with a small
transformer encoder used to produce vectors for each of the trajectories.
This encoder only consists of two self-attention layers that encode the object trajectories into vectors.

In more detail, for an object $j$, the transformer takes its ground-truth trajectory $B_j \in \mathbb{R}^{T\times 4}$ and outputs the embedding $z_{j} \in \mathbb{R}^{C}$ as the positive to be matched against the summary vector $\hat{s}_{j} \in \mathbb{R}^{C}$.
For negatives, other trajectories in the same batch as well as $n$ new ones generated 
by temporally shuffling $B_j$ encoded into $z^{shuffle}_{j}$ are used.
% The optimization objective is:\agnote{why is there hat on top of the summary vector s?}
\begin{equation}
\mathcal{L}_{aux}=-\sum_{j}\Bigg[\log{\frac{ \exp(\transpose{\hat{s}}_{j} \cdot {z}_{j} ) }
	{\sum_{k}{\exp(\transpose{\hat{s}}_{j} \cdot {z}_{k})}+ \sum_{k}{\exp(\transpose{\hat{s}}_{j} \cdot {z}^{shuffle}_{k})} } }\Bigg] 
\label{eq:aux}
\end{equation}

\noindent\textbf{Final objective.}
We use two MLPs as classifiers for the CLS vectors from visual and trajectory backbones and the Object Learner. 
The first classifier $f_{1}(.)$ is applied to concatenated $C_{visual}$ and $C_{traj}$ CLS vectors, 
and the second,  $f_{2}(.)$,  is applied to the CLS vector $C_{obj}$ from the Object Learner. 
The total loss is the sum of the cross-entropy loss for the two classifiers and the auxiliary loss:
% \agnote{Use some other symbol rather than `+' for concatenation?}
\begin{equation}
\mathcal{L}_{\text{total}} = \mathcal{L}_{CE}(f_1 (C_{visual};C_{traj}),gt) + \mathcal{L}_{CE}(f_2 (C_{obj}),gt) + \mathcal{L}_{aux},
\label{eq:loss}
\end{equation}
The final class prediction is obtained by averaging the class probabilities from
the two classifiers.\\

\noindent\textbf{Discussion: Object Learner and other fusion modules.}
Prior fusion methods can be categorized into three main types: (a)~RoI-Pooling 
based methods like STRG~\cite{strg}, where visual features
are pooled using boxes for the downstream tasks;  (b)~Joint training
methods like ORViT~\cite{orvit} where the two modalities are encoded
jointly from early stages; and (c)~Two stream methods~\cite{stlt,sthelse}
with dual encoders for the visual and trajectory modalities, where fusion is in the last layer.  The
RoI-pooling based methods explicitly pool features inside boxes for downstream operations, omitting context outside the boxes. 
In contrast, our model allows the
queries to attend to the visual feature maps freely.  Joint training
benefits from fine-grained communication between modalities, but this may
not be as robust as the two-stream models under domain shift.  Our
method combines the two, by keeping the dual encoders for independence
and having a bridging module to link the information from their
intermediate layers. Quantitative comparisons are done in \Cref{s:exps}.

\section{Implementation details}\label{s:impl}
\noindent\textbf{Model architecture.}
We use Motionformer~\cite{mformer} as the visual encoder, operating on 16 frames of size $224{\times}224$ pixels uniformly sampled from a video;
the 3D patch size for tokenization is $2{\times}16{\times}16$. 
We use STLT~\cite{stlt} as the trajectory encoder which takes 
normalized bounding boxes from 16 frames as input. 
Our Object Learner is a Cross-Transformer with 6 layers and 8 heads. We adopt the trajectory attention introduced in~\cite{mformer} instead of the conventional joint spatio-temporal attention in the layers. 
The Classification Module has 4 self-attention layers with 6 heads. We set the number of context queries as 2 in all the datasets, and number of object queries as 6 in SomethingElse, SomethingSomething and EpicKitchens, 37 in ActionGenome.\\

\noindent\textbf{Training}
We train our models on 2 Nvidia RTX6k GPUs with the AdamW~\cite{adamw} optimizer. Due to the large model size and limited compute resources, we are not able to train the full model end-to-end with a large batch size. 
Instead, we first train the visual backbone (MotionFormer) for action classification with batch size 8, and then 
keep it frozen while we train the rest of the model with batch size 72.
More details on architecture and training are in the supplementary material.

\noindent\section{Experiments}\label{s:exps}
We conduct experiments on four datasets, namely SomethingSomething-V2, SomethingElse, Action Genome and EpicKitchens.
We first train and evaluate our model on the standard task of action recognition on these datasets,
and then test its transferability on novel tasks/settings including action recognition on unseen objects, few-shot action recognition, hand state classification, and scene-graph predicate prediction.
We first introduce the datasets and the metrics. Followed by a comparison of our method with other fusion methods, and then ablations on the design choices 
in the proposed Object Learner.
Finally, we compare with SOTA models on different tasks and analyze the results.\\
% \vspace{-5mm}
\subsection{Datasets and Metrics}\label{s:exp-data}

\noindent\textbf{SomethingSomething-V2~\cite{sthsthv2}}
is a collection of 220k labeled video clips of humans performing basic actions with objects, with 168k training videos and 24k validation videos.
It contains 174 classes, these classes are object agnostic and named after the interaction, e.g, `moving something from left to the right'. 
We use the ground-truth boxes provided in the dataset as input to our networks.\\

\noindent\textbf{Something-Else~\cite{sthelse}}
is built on the videos in SomethingSomething-V2~\cite{sthsthv2} and proposes new training/test splits for two new tasks testing for generalizability: 
compositional action recognition, and few-shot action recognition.
The compositional action recognition is designed to ensure there is no overlap in object categories between 55k training videos and 58k validation videos. 
In the few-shot setting, there are 88 base actions (112k videos) for pre-training, 86 novel classes for fine-tuning.
We use the ground-truth boxes provided in the dataset.\\

\noindent\textbf{Action Genome~\cite{ji2020action}}
 is a dataset which uses videos and action labels from Charades~\cite{charades}, 
 and decomposes actions into spatio-temporal scene graphs by annotating human, objects and their relationship in them. 
 It contains 10K videos (7K train/3k val) with 0.4M objects. 
 We use the raw frames and ground-truth boxes provided for action classification over 157 classes on this dataset. \\

\noindent\textbf{Epic-Kitchens~\cite{epickitchen}} is a large-scale egocentric dataset with 100-hour activities recorded in kitchens. 
It provides 496 videos for training and 138 videos for val, each video has detected boudning boxes from~\cite{100doh}. We use an offline tracker~\cite{bytetrack} to build association between the boxes and use them as input.
We use the detected boxes provided in the dataset as input to our networks.\\

\subsection{Ablations}\label{s:exp-ablations}

\noindent\textbf{Comparison with other fusion methods.}
For a fair comparison, we implement other fusion methods using our (\ie, the same) visual and trajectory backbones.
We choose LCF (Late Concatenation Fusion) and CACNF (Cross-Attention CentralNet Fusion) as they are the two best methods among seven in the latest work~\cite{stlt}.
We also compare against a baseline method where the class probabilities from the encoders trained independently are averaged.

  \begin{wraptable}[]{}{8cm}
    \vspace{-5mm}
    \renewcommand{\arraystretch}{1.2}
    \setlength{\tabcolsep}{8pt}
    \resizebox{0.65\textwidth}{!}{%
    \begin{tabular}{cccccc}
      \hline
    \multirow{2}{*}{\textbf{\begin{tabular}[c]{@{}c@{}}Visual\\ Enc.\end{tabular}}} & \multirow{2}{*}{\textbf{\begin{tabular}[c]{@{}c@{}}Traj. \\ Enc.\end{tabular}}} & \multicolumn{2}{c}{\textbf{Fusion}}   & \multicolumn{2}{c}{\textbf{SthSth-V2}}      \\
                                                                                       &                                                                                         & \textbf{Method} & \textbf{Parametric} & \textbf{Top1}        & \textbf{Top5}        \\ \hline
    Motionformer                                                                       & -                                                                                       & -               & -                   & 66.5                 & 90.1               \\ 
    -                                                                                  & STLT                                                                                    & -               & -                   & 57                   & 85.2                 \\ \hline
    \multirow{4}{*}{Motionformer}                                                      & \multirow{4}{*}{STLT}                                                                   & avg prob             & \xmark                   &  67.5                & 90.9                 \\
                                                                                       &                                                                                          & CACNF~\cite{stlt}           & \cmark                   & 69.7                &93.4 \\ 
                                                                                       &                                                                                          & LCF~\cite{stlt}             & \cmark                   & 73.1                 & \textbf{94.2}                \\   
                                                                                       &                                                                                         & OL(ours)  & \cmark                   & \textbf{74.0}        & \textbf{94.2}        \\ \hline
    \end{tabular}%
    }
    \vspace{-3mm}
    \caption{\textbf{Different fusion methods with the same visual backbone.} We show the performance of Motionformer and STLT alone on SthSth-V2, and compare the classification performance with different fusion methods on them, namely averaged class probabilities, LCF and CACNF and our Object Learner (OL).}
    \label{tab:fusion-ours}
    \vspace{-5mm}

    \end{wraptable}
 
We implement CACNF with the same number of cross-attention layers and attention heads as in our Object Learner. \Cref{tab:fusion-ours} summarizes the results.
The results show that the above fusion methods work better than the baseline,  
and our model achieves better results than other parametric methods. Performance for Motionformer with other trajectory encoders, or STLT with other visual encoders, has been explored in previous works~\cite{orvit,stlt,sthelse}---more comparisons are 
in~\Cref{tab:sth-sota}.\\

\begin{figure}[t]
  \centering
  \includegraphics[width=0.95\linewidth]{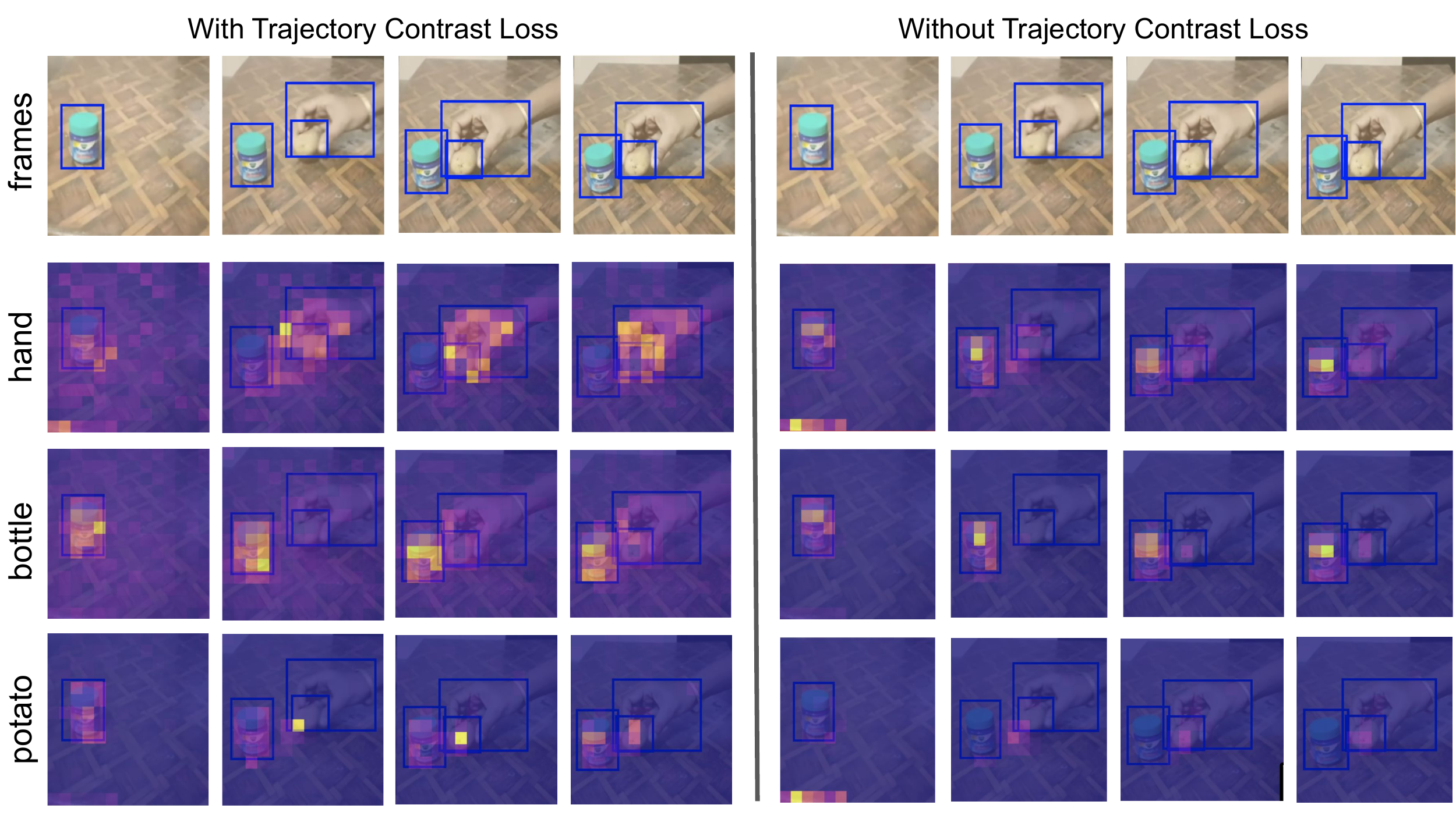}
  % \caption{\textbf{Modality fusion attention visualization.} 
  \caption{\textbf{Trajectory contrast loss induces object-centric attention.} 
  Object Learners's attention from various heads when trained with (left) and without (right) the trajectory
  contrast loss~\Cref{eq:aux}.
  In both cases, attention is on individual objects, indicating 
  objectness in the model. However, there is a notable difference: 
  without the trajectory contrast loss, there is no attention on the hands(first row). Hence, the trajectory contrast loss
  induces enhanced objectness in the object summary vectors.}
  \label{fig:attn-viz}
  % \vspace{-5mm}
\end{figure}

\noindent\textbf{Ablating trajectory contrast loss.}
We compare the performance of training with and without the trajectory contrast
loss on different transfer tasks in \Cref{tab:aux_loss}.
Having the object-level auxiliary loss (\Cref{eq:aux}) brings improvement in performance in 3 out of 4 tasks. The improvement is 1\% in hand state classification in SomethingElse, 2\% in scene-graph prediction and 1.7\% on standard classification in Action Genome.
The results show the auxiliary loss helps in both task transfer as well as 
standard action recognition.
 \Cref{fig:attn-viz} also shows the visualization of attention scores in the Object Learner -- object queries trained with auxiliary loss are more object-centric when attending to the visual frames.

\subsection{Results}\label{s:exp-results}
We present the experiment results on a wide range of tasks organized into sections
by the dataset used.
In each dataset, we first compare the results from different models on standard action recognition, 
and then introduce the transfer tasks and discuss the performance.
\subsubsection{SomethingSomething-V2 and SomethingElse} \hfill\\
\noindent\textbf{Standard Action Recognition.}
We evaluate on the regular train/val split on SomethingSomething-V2 for the performance on (seen) action recognition. 
The accuracy of our model is 5.9\% higher than ORViT and 7.1\% higher than STLT~\cite{stlt}, 
showing the advantage is not only on transfer tasks but also on standard action classification.\\
\noindent\textbf{Transfer to Unseen Objects.}
Compositional action recognition is a task in Something-Else where the actions are classified given unseen objects (\ie, objects not present in the training set). 
Thus it requires the model to learn appearance-agnostic information on the actions. 
Our object-centric model improves the visual Motionfomer by 10.9\%, and outperforms the joint encoding ORViT model by a margin of 3.9 \%, showing that keeping the trajectory encoding independent from the visual encoder can make the representations
more generalizable.\\

\begin{table}[]
    \resizebox{\textwidth}{!}{%
    \begin{tabular}{cccc|cc|cccc}
    \hline
    \multirow{3}{*}{\textbf{Model}} & \multirow{3}{*}{\textbf{\begin{tabular}[c]{@{}c@{}}Video \\ Input\end{tabular}}} & \multirow{3}{*}{\textbf{\begin{tabular}[c]{@{}c@{}}Box \\ Input\end{tabular}}} & \multirow{3}{*}{\textbf{\begin{tabular}[c]{@{}c@{}}GFLOP\end{tabular}}}  & \multicolumn{2}{c|}{\textbf{SthSth-V2}}       & \multicolumn{4}{c}{\textbf{SomethingElse}}       \\ \cline{5-10} 
                                    &                                        &                        &               & \multicolumn{2}{c|}{\textbf{Action Recognition}} & \multicolumn{2}{c}{\textbf{\begin{tabular}[c]{@{}c@{}}Compositional Action\\ (Unseen Objects) \end{tabular}}} & \multicolumn{2}{c}{\textbf{Few Shot}}       \\
                                    &                                        &                        &                & Top 1                   & Top5                 & Top 1                                               & Top 5                                              & 5 shot               & 10 shot                   \\ \hline
    I3D ~\cite{carreira2017quo}     & \cmark                                 & \xmark               &  28              & 61.7                    & 83.5                  & 46.8                                                & 72.2                                               & 21.8                 & 26.7  \\
    SlowFast,R101~\cite{slowfast}   & \cmark                                 & \xmark               & 213               & 63.1                    & 87.6                  & 45.2                                                & 73.4                                               & 22.4                 & 29.2     \\
    Motionformer~\cite{mformer}     & \cmark                                 & \xmark               &  369              & 66.5                    & 90.1       & 62.8                                                & 86.7                                               & 28.9                 & 33.8                 \\ \hline
    STIN~\cite{sthelse}             & \xmark                                 & \cmark               &  5.5               & 54.0                      & 79.6     & 51.4                                                & 79.3                                               & 24.5                 & 30.3                               \\
    SFI~\cite{sfi}                  & \xmark                                 & \cmark               &  -               & -                       & -             & 44.1                                                & 74                                                 & 24.3                 & 29.8                          \\
    STLT~\cite{stlt}                & \xmark                                 & \cmark               &  4                 & 57.0                      & 85.2       & 59.0                                                  & 86                                                 & 31.4                 & 38.6                         \\ \hline
    STIN+I3D~\cite{sthelse}         & \cmark                                 & \cmark               &  33.5             & -                       & -               & 54.6                                                & 79.4                                               & 28.1                 & 33.6                     \\
    STIN,I3D~\cite{sthelse}         & \cmark                                 & \cmark               & 33.5          & -                       & -                 & 58.1                                                & 83.2                                               & 34.0                   & 40.6                     \\
    SFI~\cite{sfi}                  & \cmark                                 & \cmark               & -             & -                       & -                   & 61.0                                                  & 86.5                                               & 35.3                 & 41.7                     \\
    R3D,STLT(CACNF)~\cite{stlt}     & \cmark                                 & \cmark               & 48             & 66.8                    & 90.6               & 67.1                                                & 90.4                                               & 37.1                 & 45.5        \\
    ORViT~\cite{orvit}              & \cmark                                 & \cmark               & 405              & 73.8                    & 90.5               & 69.7                                                & 91                                                 & 33.3                 & 40.2                     \\ \hline
    Motionformer+STLT(baseline)               & \cmark                                 & \cmark                & 373              & 72.8                    & 94.1                                                                & 72.0                                                  & 92.3         & 38.9                 &  44.6                                \\
    % Ours - Object Learner Only                      & \cmark                                      & \cmark               &                     & 72.1                    & 93.8         & 71.0                                                  & 92.4                                               & 38.8                  & 44.9                                \\
    Motionformer+STLT+OL(Ours)      & \cmark                                      & \cmark                   & 383.3               & \textbf{74.0}           & \textbf{94.2}      & \textbf{73.6}                           & \textbf{93.5}          & \textbf{40.0}     & \textbf{45.7}                                                 \\ \hline
    \end{tabular}%
    }
    \vspace{2mm}
    \caption{\textbf{Comparison with SOTA models on Something-Else and SomethingSomething-V2.} We report top1 and top5 accuracy on three action classification tasks, including compositional and few-shot action recognition on Somehing-Else, 
    and action recognition on SomethingSomething-V2. From top to bottom: we show the performance of SOTA visual models, trajectory models, and the models which takes both modalities as input. 
    In the last section we list the classification performance from the backbone baseline without an Object Learner(OL) and our model with an Object Learner(OL), Our model outperforms other methods by a clear margin on all the tasks.}
    \label{tab:sth-sota}
    \vspace{-4mm}

\end{table}

\noindent\textbf{Data-efficiency: Few-shot Action Recognition.}
We follow the experiment settings in \cite{sthelse} to freeze all the parameters except the classifiers in 5-shot and 10-shot experiments on SomethingElse.
Again, models that are using both visual and trajectory modalities have an obvious advantage over visual only ones.
The performance boost is more obvious in a low data regime, with a 4.7\% and 0.3\% improvement over R3D, STLT in 5-shot and 10-shot respectively.
It's worth noting that while the raw classification results from the backbone and Object Learner classifiers only have a 0.1\% difference, averaging the two together gives more than 1\% improvement. 
It suggests that our Object Learner has captured complementary information through combining the two streams.\\

\noindent\textbf{Transfer to Hand State Classification.}
We further evaluate the object-level representations (pre-trained with standard
action recognition) on hand contact state classification using a linear probe.
We extract hand state labels using a pre-trained object-hand detector from \cite{100doh} as ground truth, and design a 3-way classification task on SomethingElse.  Specifically, the three classes are `no hand contact', `one hand contact' (one hand contacts with object) or `two hands contact' (both hands contact with object).
In our experiments, we average-pool the object summary vectors, train a linear classifier on the training set, and test on the validation set.
We conduct the linear probe on summaries trained with and without the auxiliary loss in \Cref{eq:aux}, and also on the baseline backbone classifier. Video-level top-1 accuracy and class-level top-1 accuracy are reported in \Cref{tab:aux_loss}.
Our model is better than the baseline by 11.4 \% in per-video accuracy and 26.4 \% in per-class accuracy. Object summaries trained with auxiliary losses on trajectories outperform the one without by about 1\%. 
\\
\begin{table}[]
    \centering
    \vspace{-10mm}
    \renewcommand{\arraystretch}{1.3}
    \setlength{\tabcolsep}{3pt}
    \resizebox{0.98\textwidth}{!}{%
    \begin{tabular}{c|c|cccc|ccc}
    \hline
    \multirow{3}{*}{\textbf{Method}} & \multirow{3}{*}{\textbf{Aux Loss}} & \multicolumn{4}{c|}{\textbf{Something-Else}} &  \multicolumn{3}{c}{\textbf{Action Genome}}                                                                                                                                            \\ \cline{3-9} 
                                                    &                                    & \multicolumn{2}{c|}{\textbf{\begin{tabular}[c]{@{}c@{}}Compositional Action\end{tabular}}} & \multicolumn{2}{c|}{\textbf{\begin{tabular}[c]{@{}c@{}}Hand Contact State\end{tabular}}} &\multicolumn{1}{c|}{\textbf{\begin{tabular}[c]{@{}c@{}}Action\end{tabular}}}  &\multicolumn{2}{c}{\textbf{\begin{tabular}[c]{@{}c@{}}Predicate\end{tabular}}} \\
                                                    &                                    & \textbf{Top1}                       & \multicolumn{1}{c|}{\textbf{Top5}}                      & \textbf{Per-video}                                  & \textbf{Per-class}    & \multicolumn{1}{c|}{\textbf{mAP}}   & \textbf{R@10}              & \textbf{R@20}                                 \\ \hline
    ORViT                                              & -                                  & 69.7                                  & \multicolumn{1}{c|}{91.0}                               &   70.2                                            &       66.0      & \multicolumn{1}{c|}{-}     & -    & -                              \\ \hline
    MFormer+STLT(baseline)                           & -                                  & 72                                  & \multicolumn{1}{c|}{93.2}                               &   66.8                                            &       43.3          & \multicolumn{1}{c|}{66.0}    &78.3   &83.5                           \\
    MFormer+STLT+OL(ours)                   & \xmark                                  & 73.5                                & \multicolumn{1}{c|}{\textbf{93.5}}                              &   77.5                                            &    68.5         & \multicolumn{1}{c|}{64.9}    &78.9 & 83.8                               \\
    MFormer+STLT+OL(ours)                   & \cmark                                 & \textbf{73.6}                       & \multicolumn{1}{c|}{\textbf{93.5}}                      &   \textbf{78.2}                                         &   \textbf{69.7}     & \multicolumn{1}{c|}{\textbf{66.6}}   &\textbf{80.9} &\textbf{85.4}                                         \\ \hline
    \end{tabular}%
    }
    \vspace{1mm}
    \caption{\textbf{Ablate auxiliary loss and Object Learner (OL) on compositional action, hand state classification and predicate prediction.} 
    We show linear probe results on the backbone CLS token, and our object summary vectors trained with and without auxiliary loss.}
    \label{tab:aux_loss}
    \vspace{-10mm}
    \end{table}

\renewcommand{\arraystretch}{1.3}
\setlength{\tabcolsep}{1pt}

    \begin{table}[]
        \centering
        \vspace{-5mm}

        \resizebox{0.95\textwidth}{!}{%
        \begin{tabular}{cc|ccc|ccc|ccc}
        \hline
        \multirow{2}{*}{\textbf{Methods}}                   & \multirow{2}{*}{\textbf{Box input}} & \multicolumn{3}{c|}{\textbf{Overall}}           & \multicolumn{3}{c|}{\textbf{Tail Classes}}      & \multicolumn{3}{c}{\textbf{Unseen Kitchens}}     \\ \cline{3-11} 
                                                            &                                     & \textbf{Action} & \textbf{Verb} & \textbf{Noun} & \textbf{Action} & \textbf{Verb} & \textbf{Noun} & \textbf{Action} & \textbf{Verb} & \textbf{Noun} \\ \hline
        SlowFast~\cite{slowfast}      & N                                   & 38.5            & 65.5          & 50.0          & 18.8            & 36.2          & 23.3          & 29.7            & 56.4          & 41.5          \\
        ViViT-L~\cite{arnab2021vivit} & N                                   & 44.0            & 66.4          & 56.8          & -               & -             & -             & -               & -             & -             \\
        MFormer~\cite{mformer}        & N                                   & 43.1            & 66.7          & 56.5          & -               & -             & -             & -               & -             & -             \\
        MFormer-HR~\cite{mformer}     & N                                   & 44.5            & 67.0          & 58.5          & 19.7            & 34.2          & 28.4          & 34.8            & 58.0          & 46.6          \\ \hline
        MFormer-HR+STRG                                     & Y                                   & 42.5            & 65.8          & 55.4          & -               & -             & -             & -               & -             & -             \\
        MFormer-HR+STRG+STIN                                & Y                                   & 44.1            & 66.9          & 57.8          & 24.7               & \textbf{39.9}             & 34.4             & 34.8               & 59.5             & 48.1             \\
        MFormer-HR-ORVIT~\cite{orvit} & Y                                   & 45.7            & 68.5          & 57.9          & -               & -             & -             & -               & -             & -             \\ \hline
        MFormer-HR+STLT(baseline)                                & Y                                   & 44.6            & 67.4          & 58.8          & 23.3            & 38.5          & 34.1          & 35.1            & \textbf{59.7}          & \textbf{49.6} \\
        MFormer-HR+STLT+OL(ours)                            & Y                                   & \textbf{46.3}   & \textbf{68.7} & \textbf{59.4} & \textbf{25.7}   & \textbf{39.9} & \textbf{35.3} & \textbf{35.4}   & \textbf{59.7} & 48.3          \\ \hline
        \end{tabular}%
        }
        \vspace{1mm}
        \caption{\textbf{Action Classification results on Epic-Kitchens.} Our model achieves the best results compared to other methods using the same backbone. MFormer uses 224x224 resolution input and MFormer-HR uses 336x336 resolution input. }
        \label{tab:ek}
        \vspace{-10mm}
        \end{table}

\subsubsection{Epic-Kitchens} \hfill\\
\noindent\textbf{Standard Action Recognition.}
\Cref{tab:ek} shows the results of action recognition in Epic-Kitchens,  
With the Object-Learner, our model is 0.6-4.0\% more accurate  in action prediction than other methods that use both visual stream and trajectory stream as input, and 1.7\% more accurate than   
the Late Concatenation Fusion (LCF) method without an Object-Learner.
\\

\noindent\textbf{Classification on Tail Classes and Unseen Kitchens.}
In \Cref{tab:ek} we also present the classification results on tail classes and videos from unseen kitchens. In average, Object-centric models are better than visual-only models by 4.8\% on tail actions, and by 0.4\% on unseen kitchens. 
Among all the models with objectness, our model with Object Learner achieves the best action classification accuracy on both tail classes and unseen kitchens.

\renewcommand{\arraystretch}{1.1}
\begin{table}[]
   \centering
   \resizebox{0.95\textwidth}{!}{%
   \begin{tabular}{ccccccccc}
   \hline
   \multirow{2}{*}{\textbf{Backbone}}           & \multirow{2}{*}{\textbf{Method}}               & \multirow{2}{*}{\textbf{Boxes}} & \multirow{2}{*}{\textbf{SG}} & \multirow{2}{*}{\textbf{Aux loss}} & \multirow{2}{*}{\textbf{\# Frames}} & \textbf{Action CLS.} & \multicolumn{2}{c}{\textbf{Predicate Pred.}} \\
                                                &                                                &                                 &                              &                                    &                                     & \textbf{mAP}                   & \textbf{R@10}           & \textbf{R@20}           \\ \hline
   I3D~\cite{stlt}        & Avgpool                                        & N                               & N                            & -                                  & 32                                  & 33.5                           & -                       & -                       \\
   MFormer~\cite{mformer} & CLS token                                      & N                               & N                            & -                                  & 16                                  & 36.5                           & 76.4                    & 82.6                    \\ \hline
   STLT~\cite{stlt}       & CLS token                                      & Y                               & N                            & -                                  & 16                                  & 56.7                           & 79.0                    & 84.1                    \\
   STLT~\cite{stlt}       & CLS token                                      & Y                               & N                            & -                                  & 32                                  & 60.0                           & -                       & -                       \\ \hline
   I3D+STLT~\cite{stlt} & CACNF~\cite{stlt}        & Y                               & Y                            & -                                  & 32                                  & 61.6                           & -                       & -                       \\
   MFormer+STLT                               & CACNF~\cite{stlt}        & Y                               & N                            & -                                  & 16                                  & 64.2                           & -                       & -                       \\\hline
   R101-I3D-NL~\cite{lfb} & SGFB~\cite{ji2020action} & Y                               & Y                            & -                                  & 32+                                 & 60.3                           & -                       & -                       \\ \hline
   MFormer+STLT                               & LCF(baseline)         & Y                               & N                            & -                                  & 16                                  & 66.0                           & 78.3                    & 83.5                    \\ 

   MFormer+STLT                                & OL(ours)                           & Y                               & N                            & N                                  & 16                                  & 64.9                           & 78.9                    & 83.8                    \\
   MFormer+STLT                                & OL(ours)                           & Y                               & N                            & Y                                  & 16                                  & \textbf{66.6}                  & \textbf{80.9}           & \textbf{85.4}           \\ \hline
   \end{tabular}%
   }
   \vspace{1mm}

   \caption{\textbf{Action recognition and human-object predicate prediction results on Action Genome.} In action classification, our model outperforms others with the same frame and boxes input, and even SGFB with scene graph input. When linear-probing the models's output for predicate prediction, 
   our Object Learner fuses the visual and trajecory streams in an efficient way and is 2.6\% higher than baseline LCF in recall@10. We also show the object-centric representations learned with the auxiliary loss is better then object-aware ones learned without the auxiliary loss in both tasks. }
   \label{tab:action-genome}
   \vspace{-8mm}
   \end{table}

\subsubsection{Action Genome}
\noindent\textbf{Standard Action Recognition.}
In Action Genome, each action clip is labelled with object bounding boxes and their categories. 
We follow the experiment settings in \cite{stlt}, train and evaluate our model with RGB frames and ground truth trajectory as input.  
\Cref{tab:action-genome} shows the classification results. 
By using our Object Learner trained with auxiliary loss, we achieve the best result 66.6\% mAP, outperforming other fusion methods using the same backbone.  
We also compare to SGFB~\cite{ji2020action}, which uses scene graphs as input, our model is better by 6.3\% without access to the relationship between objects.\\

\noindent\textbf{Transfer to scene graph predicate prediction.}
We transfer the trained model on action classification to scene graph predicate prediction by linear probing. 
% Again, we freeze the model trained on action recognition and only learn a linear classifier for predicate prediction on the training set.
In this task, the model has to predict the predicate (relationship) between human and object when the bounding boxes and categories are known. 
Given the object id, we concat one-hot object id vectors with the classification vector from the frozen models, and train a linear classifier to predict the predicate.
As shown in \Cref{tab:action-genome}, the result from object summaries trained with the auxiliary loss is 2.6\% higher than linear probing the concatenated CLS tokens (LCF) from two backbones, 
and 2.0\% higher than the one trained without auxiliary loss.

\section{Conclusion} \label{s:conc}
We set out to evaluate whether objectness in video representations can aid
visual task transfer.
To this end, we have developed an object-centric video model which 
fuses the visual stream with object trajectories (bounding-boxes) in a novel
transformer based architecture.
We indeed find that the object-centric representations learned by our model are
more transferrable to novel tasks and settings in video recognition using a simple
linear probe, \ie, they outperform both prior object-agnostic and object-centric
representations on a comprehensive suite of transfer tasks. Furthermore, they 
also set a new record on the standard task of action classification on a number
of benchmarks.
This work uses a very coarse geometric representation of objects, \ie,
 bounding-boxes, for inducing object awareness in visual representations;  in the future
more spatially precise/physically-grounded representations, \eg, segmentation 
masks, 3D shape could further enhance the transferability. \\

\noindent\textbf{Acknowledgements.} This research is funded by
 a Google-DeepMind Graduate Scholarship, a Royal Society Research Professorship, and 
EPSRC Programme Grant
VisualAI EP/T028572/1.
% segmentation
% 3d shape
% more attributes, even better for transfer.

% While we establish that object-centric representations are more
% transferrable and generalize better to novel tasks, how to induce such objectness
% in visual representations without relying on human-annotated object cues remains 
% an open question.

% input to achieve good transferability across 
% datasets and tasks. With a newly-proposed Object Learner module, we build effective fusion the two modalities resulting in repurposable object-aware representations.
% We evaluate our model on four action recognition benchmarks.
% The evaluations show that the new object-aware model achieves state-of-the-art performance not only on the traditional action recognition, 
% but also on compositional action recognition with unseen objects, and in a few-shot setting on novel action categories.
% Furthermore, these representations can be effectively be adapted for other object-oriented downstream task like hand contact state classification and predicate prediction with a simple linear regressor.
\bibliographystyle{splncs}
\bibliography{bib/shortstrings,bib/vgg_local,bib/vgg_other,bib/refs}
%******************
\newpage
\appendix
\renewcommand{\contentsname}{Appendix}
\vspace{12mm}
{%
\let\clearpage\relax
\hypersetup{linkcolor=black}
\tableofcontents
}

\clearpage
% ******************
\section{Architecture}
\subsection{Visual backbone}
We use Motionformer~\cite{mformer} as the visual backbone, it takes a sequence of video frames $I \in \mathbb{R}^{T \times H\times W\times 3}$, patchifies these into 3D patches of size $(2\times 16\times 16\times 3)$ each, and then encodes them.
It has 12 self-attention layer with 12 heads each, and outputs feature maps $V\in \mathbb{R}^{T'\times (H'W')\times C}$, where $C=768$. All the hyper-parameters are the same as the ones used in \cite{mformer} on SomethingSomething-V2.

\subsection{Trajectory backbone}
The  trajectory backbone consists of a box embedding module and a Spatial Temporal Layout model (STLT) of~\cite{stlt}. 
It takes a sequence of bounding boxes of objects as input and outputs spatial and temporal layout embeddings. 

The input bounding boxes ${B^t} = (b^t_1, b^t_2, b^t_3,
\hdots,b^t_o)$ of $\mathcal{O}$ number of objects in a given frame are in the 
format $[x_1,y_1,x_2,y_2]$, they are first projected into box embeddings 
${\Phi} \in \mathbb{R}^{O \times T \times C} $ through an MLP.
Learnable object-ID embeddings $\mathcal{D}=\{d_j\}_{j=1}^{O}$ are added to the box embeddings to 
obtain ${\Phi_{in}}$, which serve as an input to the STLT .

\begin{figure}[t]
  \centering
  \includegraphics[width=0.95\linewidth]{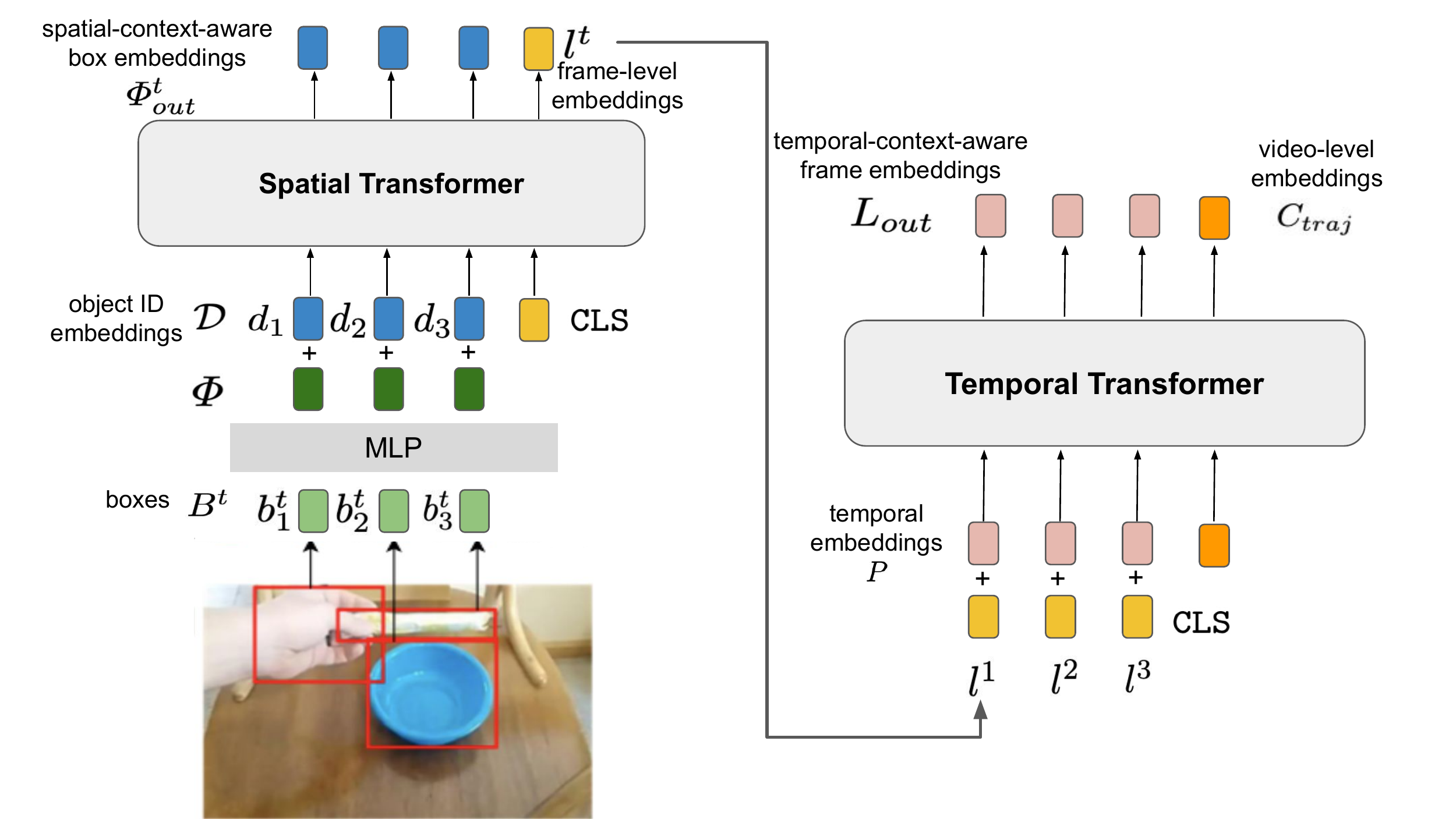}
%   \vspace{-5mm}
  \caption{\textbf{Architecture of STLT~\cite{stlt}.} \textbf{Left: Spatial Transformer:} 
  It takes a set of object bounding boxes $B^t$ at different frames and encodes their spatial layout independently into spatial-context-aware embeddings $\phi_{out}^t$,
   a special class embedding is used to aggregate per-frame information into $l^t$.
  \textbf{Right: Temporal Transformer:} It takes the frame-level output $l^t$ from the Spatial Transformer as input, 
  and encodes them along the temporal dimension into temporal-context-aware embeddings $L_out$ . 
  Another special class embedding is concatenated to the temporal features for video-level representation $C_{traj}$. 
  $C_{traj}$ is used in computing the final loss in \cref{eq:loss}, and spatio-temporal embeddings $\phi_{out}$ and $L_{out}$ are used later as input to Object Learner.}  
%   \vspace{-6mm}
  \label{fig:stlt}
\end{figure}

STLT consists of two self-attention transformers, the Spatial Transformer and the Temporal Transformer.  
An overview of its archiecture is shown in \Cref{fig:stlt}.
The Spatial Transformer processes the boxes at each frame separately. 
In each frame, it takes a learnable \texttt{CLS} token and box embeddings ${\Phi_{in}^t} \in \mathbb{R}^{O \times C}$ as input into the self-attention layers,
and output a frame-level representation ${l^t} \in \mathbb{R}^{1\times C}$ and spatial-context-aware box embeddings ${\Phi_{out}^t} \in  \mathbb{R}^{O \times C}$.

The Temporal Transformer encodes trajectory information between frames, it applys self-attenion on the frame-level embeddings ${L^t} = (l^1, l^2, \hdots,l^T)$ from the Spatial Transformer with another learnable \texttt{CLS} token. 
At the output,  we will have temporal-context-aware frame embeddings ${L_{out}} \in \mathbb{R}^{T \times C}$  and a video-level representation ${C_{traj}} \in \mathbb{R}^{1\times C}$. 
$C_{traj}$ is later used to compute the classification loss, while ${L_{out}} \in \mathbb{R}^{T \times 1 \times C}$ is concatenated with the 
${\Phi_{out}} \in \mathbb{R}^{T \times O \times C}$ from the Spatial Transformer as the final trajectory embeddings 
$G \in \mathbb{R}^{T\times (O+1)\times C}$, 
$G$ is then downsampled over the temporal dimension from $T$ to $T'$, so that it can be concatenated with the visual feature map $V \in \mathbb{R}^{T'\times H'W'\times C}$ from the 6th layer of the visual backbone.
The concatenated results serve as the keys and values of the Object Learner.
% For one video, the keys and values are $K/V \in \mathbb{R}^{T'\times (H'W'+O+1)\times C}$, where $C=768$.

\subsection{Object Learner} 
The Object Learner consists of 4 cross-attention layers, each with 4 heads. The input feature dimension is 512 and the feed-forward dimension is 2048.
Keys, values and the object-ID embeddings $\mathcal{D}$ from the backbones are linearly projected to dimension 512.
Queries into the Object Learner are the sum of the projected object-ID embeddings $\mathcal{D'}=\{d'_i\}_{i=1}^{O}$ and a set of learnable embeddings $\mathcal{Q}=\{q_i\}_{i=1}^{O+K}$, 
where $O$ is the number of object queries and $K$ is the number of context queries. We set $O=5$, $K=3$ in the experiments.
And the keys and values are the concatenation of spatio-temporal visual features $V \in \mathbb{R}^{T'\times H'W'\times C}$ and the temproally downsampled trajectory features $G \in \mathbb{R}^{T'\times (O+1)\times C}$.

% Features from the backbones are first projected from 768 to 512 before entering the Object Learner by a linear layer, the feed-forward dimension in the layers is 2048.
We use trajectory attention mechanism~\cite{mformer} in cross-attention layers to replace joint spatio-temporal attention. 
For each object query $q_{i}$, we first compute its attention scores along the spatio-temporal dimension,
and use the scores for weighted pooling on spatial dimension only:

\begin{equation}
    \mathbf{a}_{ist} =
    \frac
    {\exp\langle \mathbf{q}_{i}, \mathbf{k}_{st}\rangle}
    {\sum_{s't'}\exp \langle \mathbf{q}_{i}, \mathbf{k}_{s't'}\rangle} ,
\end{equation}

\begin{equation}
    \mathbf{\tilde{y}}_{it} =
    \sum_{s}
    \mathbf{v}_{st} \cdot \mathbf{a}_{ist} ,
\end{equation}
where $\tilde{y}_{it}$ is the spatially aggregated token at time $t$ given $q_{i}$, which is also referred as the `trajectory token' at time $t$.
Once the trajectories $\tilde{Y}_{i}$ are computed, they are further pooled across time to extract intra-frame
information/connections. To do so, the trajectory tokens are projected to a new set of keys and values, and the query is projected again to a new set of temporal queries:
\begin{equation}
    \tilde{\mathbf{q}}_{i} = \tilde{\mathbf{W}}_q\,
    \mathbf{q}_{i},
    ~~~
    \tilde{\mathbf{k}}_{it} = \tilde{\mathbf{W}}_k\,
    \tilde{\mathbf{y}}_{it},
    ~~~
    \tilde{\mathbf{v}}_{it} = \tilde{\mathbf{W}}_v\,
    \tilde{\mathbf{y}}_{it}.
\end{equation}
The new query is used to pool across the new time (trajectory) dimension by applying 1D cross-attention:
\begin{equation}
    \mathbf{y}_{i}
    =
    \sum_{t}
    \tilde{\mathbf{v}}_{it} \cdot
    \frac
    {\exp\langle \tilde{\mathbf{q}}_{i}, \tilde{\mathbf{k}}_{it}\rangle}
    {\sum_{t'}\exp \langle \tilde{\mathbf{q}}_{i}, \tilde{\mathbf{k}}_{it'}\rangle}.
\end{equation}

\subsection{Classification Module \& classifier}
The Classification Module is made up of 2 self-attention layers, each with 4 heads. The input dimension of features is 512 and the feed-forward dimension 2048. A learnable CLS token is concatenated to the input features, the output of which is then fed into a downstream classifier for final classification.
The downstream classifier is an MLP with two linear layers and a tanh activation between them.

\section{Hand Contact State Classification}
\begin{figure}[]
    \vspace{-5mm}
  \centering
  \includegraphics[width=0.95\linewidth]{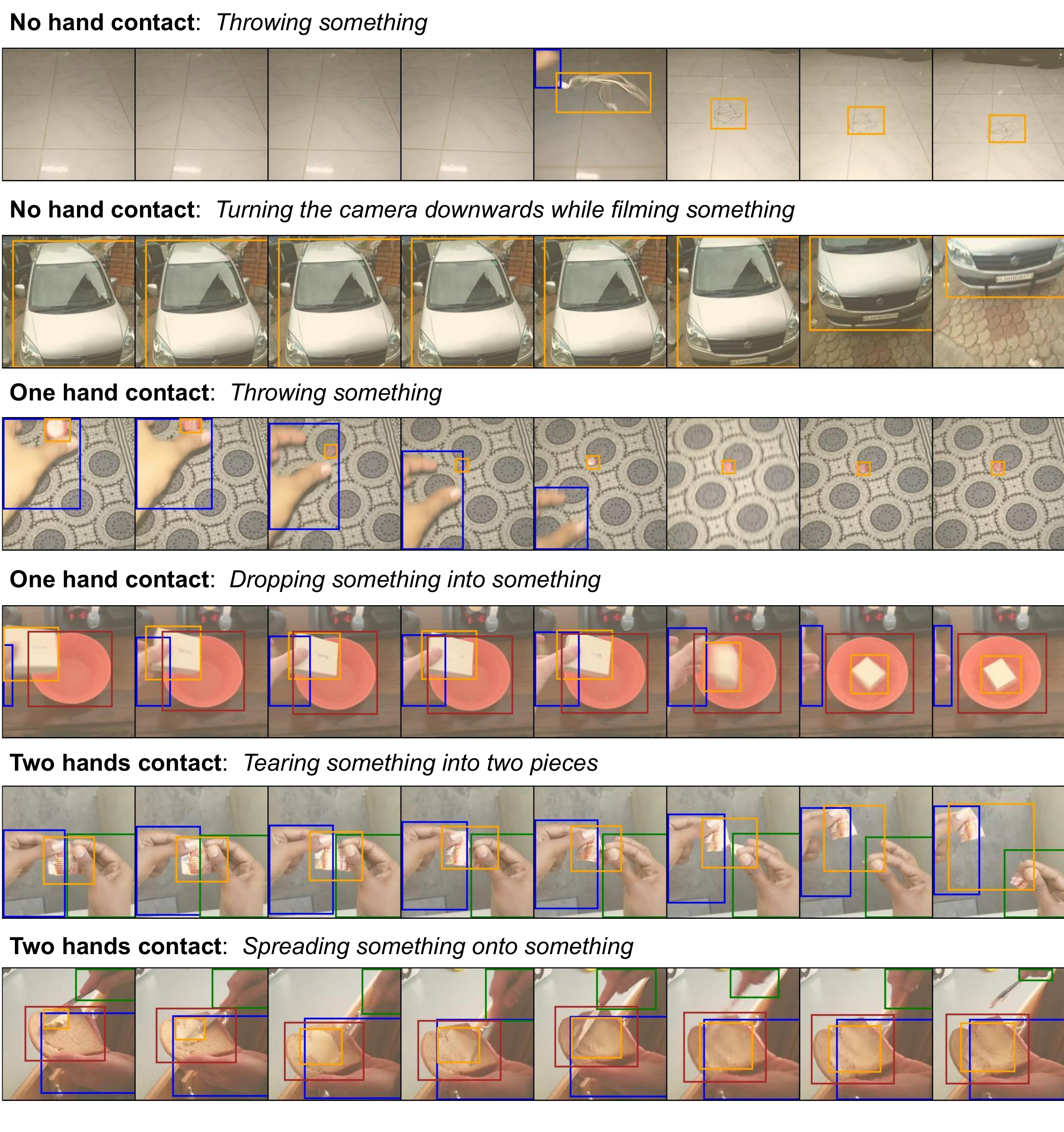}
  \vspace{-5mm}
  \caption{\textbf{Visualization of samples from the three classes in hand contact state classification.} 
Above each video sample, we show its hand contact state class together with its action class. 
  Ground-truth bounding boxes of hands and objects are plotted in the frames, 
  with \textcolor{blue}{blue} and \textcolor{teal}{green} boxes on hands. \textcolor{Dandelion}{yellow} and \textcolor{red}{red} boxes on objects. }
  \vspace{-6mm}
  \label{figs:hand-state-viz}
\end{figure}

We use the training and validation split in SomethingElse~\cite{sthelse} for hand contact state classification.
To generate `ground truth' contact state labels, we use a pre-trained object-hand state detector from~\cite{100doh}. The detector predicts 5 hand contact states, namely `no contact', `self contact', `other person contact', `portable object contact' and `stationary object contact' (e.g., furniture).  
It labels 85\% of the frames in SomethingElse as `portable object contact' and the rest as other types of contact. 
Instead of doing classification on a very unbalanced  contact state, we design a 3-way classification task by categorizing the videos into the following classes: 
\begin{enumerate}
    \item No hand contact: there is no hand in the video, or there are hands in the frames but they are not in contact with any object.
    \item One hand contact: There are one or two hands in the video, only one hand is in contact with objects.
    \item Two hands contact: There are two hands in the video, they in contact with the same or different objects.
\end{enumerate}
To do this video-level categorization, frame with the largest number of hands detected are used from each video, 
we check whether these hands are labelled as `no contact' to decide which class the video falls in.
The class distribution is shown in  \Cref{{tab:contact_state}}.  Examples of the 3 classes are visualized in \Cref{figs:hand-state-viz}.
% \item one hand contact' (one hand contacts with object) or `two hands contact' (both hands contact with object) using the labels predicted.
% The following table shows the statics of the 3 classes:

\renewcommand{\arraystretch}{1.4}

\begin{table}[]
  \centering
  \vspace{-2mm}
  \resizebox{0.7\textwidth}{!}{%
  \begin{tabular}{c|c|c|c}
  \hline
  Class    & no contact & one hand contact & two hands contact \\ \hline
  \%videos & 9\%        & 31\%             & 59\%              \\ \hline
  \end{tabular}%
  }
  \vspace{2mm}
  \caption{\textbf{Distribution of classes in hand contract state classification in SomethingElse.} We use the labels provided by a pre-trained hand state detector~\cite{100doh}, and categorize the videos into 3 classes: `no hand contact', `one hand contact' and `two hands contact'.}
  \label{tab:contact_state}
  \vspace{-10mm}
\end{table}
\clearpage 

\section{Human-Object Predicate Prediction}
Given the bounding box and category of an object, the model is tasked to predict the predicate between human and this object. 
Action Genome~\cite{ji2020action} has annotations in 37 categories (36 objects + 1 human). There are 25 human-object relationships (aka. predicate), including 3 attention relationships, 6 spatial relationships 
and 16 contact relationships. There can be more than one relationships between a person and an object, thus the performance is measured in terms of recall. \\
When we linear probe our baselines Motionformer and Motionformer+STLT, we use the CLS tokens from the backbones, 
and concatenate them with a one-hot object-id, indicating which object we want to predict the predicate.
When linear probing our model with an Object Learner, we use the same approach except that we are also 
concatenate the CLS tokens and object-ids with the addtiontal object-centric representations of the given object.
We train a 25-way linear classifier on the the concatenated vector to predict the predicate classes, using a binary cross-entropy loss.

\section{Ablations}
\subsection{Number of context queries}

We ablate the number of context queries in our Object Learner. 
In \Cref{tab:contextq} we show the classification performance with $\{0,3,6,9\}$ 
context queries on SomethingElse. The top-1 accuracy increases by only 0.2\% as the
number of context queries goes from 0 to 9. The small impact might be due to the fact that action recognition in the dataset we use only depends on  two or three key objects.

% \begin{wraptable}[]{}{6cm}
%     \centering
%     % \renewcommand{\arraystretch}{1.3}
%     \setlength{\tabcolsep}{8pt}
%     \resizebox{0.7\textwidth}{!}{%
%     \begin{tabular}{c|cccc}
%     \hline
%     \# \textbf{context queries} & \textbf{0} & \textbf{3}    & \textbf{6} & \textbf{9} \\ \hline
%     Top 1              & 73.5  & 73.6 & 73.7  & 73.7  \\
%     Top 5              &  93.5 & 93.5 &  93.5 & 94.6 \\ \hline
%     \end{tabular}%
%     }
%     \vspace{2mm}
%     \caption{\textbf{Ablation on number of context queries in Object Learner on SomethingElse.}}
%     \label{tab:contextq}
% \end{table}

\subsection{Ablation on choice of input layer from the visual backbone}
We ablate the performance of models with the Object Learner reading from different layers in the visual backbone. We tried layers 6,8,12 from a Motionformer with 12 layers in total.
\Cref{tab:visual-layers} shows the results. While the accuracy of direct class predictions from our Object Learner does not differ too much ($\pm 0.2\%$), the input visual layer has a big influence on the combined results from Object-Learner and CLS token, where we averge the probability prediction from Object Learner and CLS token.
The improvement on the averaged Top1 is 1.6\% when using layer 6, and -0.1\% when using layer 12. The monotonic drop with increase in depth suggests that earlier layer fusion is necessary for complementary results to our Object Learner.

\begin{table}[!htb]
    \hspace{2mm}
    \begin{minipage}{.42\linewidth}
        \centering
        \setlength{\tabcolsep}{3pt}

        \begin{tabular}{c|cccc}
            \hline
            \textbf{\begin{tabular}[c]{@{}c@{}}\#context \\ queries\end{tabular}} & \textbf{0}  & \textbf{3} & \textbf{6} & \textbf{9} \\ \hline
            Top 1                                                                   & 73.5    & 73.6       & 73.7       & \textbf{73.7}       \\
            Top 5                                                                   &  93.5   & 93.5       & 93.5       & \textbf{93.6}       \\ \hline
            \end{tabular}%
        \vspace{2mm}
        \caption{\textbf{Ablation on number of context queries in Object Learner.} We evaluate the compositional action recognition performance on SomethingElse by using different number of context queries. }
        \label{tab:contextq}
    \end{minipage}%
    \hspace{4mm}
    \begin{minipage}{.55\linewidth}
        \centering
        \setlength{\tabcolsep}{3pt}
        \renewcommand{\arraystretch}{1.2}
        \begin{tabular}{c|ccc}
            \hline
            \textbf{\begin{tabular}[c]{@{}c@{}} Input \\ Visual Layer \end{tabular}} & \textbf{\begin{tabular}[c]{@{}c@{}}OL \\ Top1\end{tabular}} & \textbf{\begin{tabular}[c]{@{}c@{}}Backbone \\ Top1\end{tabular}}& \multicolumn{1}{c}{\textbf{\begin{tabular}[c]{@{}c@{}} Avg \\ Top1\end{tabular}}} \\ \hline
            6                                                                                       & 71.0                                                                   &             72.0                  & \textbf{73.6}                                            \\
            8                                                                                    & \textbf{71.3}                                                                      &             \textbf{72.3}                  & 73.1                                        \\ 
            12                                                                                    & 70.9                                                                     &              72.1                 & 72.0                                           \\\hline
            \end{tabular}%
            \vspace{2mm}
            \caption{\textbf{Ablation on the Visual Input in Object Learner.} We evaluate the performance of our model on compositional action recognition (unseen objects) with the Object Learner extracting features from layer 6,8,12 in the visual backbone of depth 12. Results show reading from the sixth layer yield best performance.}
            \label{tab:visual-layers}
    \end{minipage} 
\end{table}

\subsection{Ablation on the depth and width of Object Learner.}
We evalute the performance of our model using an Object Learner with a varying number of layers and heads.
\Cref{tab:ol-depth} shows the results ranging from 4 layers to 8 layers, and from 4 heads to 8 heads.  
Doubling the size of model only leads to 0.2\% increase in top1 accuracy and 0.3\% increse in top5 accuracy. 
It shows that learning good object-centric representations from small number of objects (within 5) does not require a very deep and wide Object Learner. 
\begin{table}[]
    \centering
    \setlength{\tabcolsep}{5pt}
    \resizebox{0.6\textwidth}{!}{%
    \begin{tabular}{cc|ccc}
    \hline
    \multicolumn{1}{l}{\textbf{\#Layers}} & \textbf{\# Heads} & \textbf{GFLOP} & \textbf{Top1} & \textbf{Top5} \\ \hline
    4                                     & 4                 & 382             & 73.6          & 93.5          \\ \hline
    8                                     & 4                 & 384.5           & 73.7          & 93.8          \\ \hline
    8                                     & 8                 & 384.5          & \textbf{73.8}          & \textbf{93.8}          \\ \hline
    \end{tabular}%
    }
    \vspace{1mm}
    \caption{\textbf{Ablation on the depth and width of Object Learner on SomethingElse.} We evalute the performance of our model using Object Learners with 4 layers and 4 heads, 8 layers and 4 heads, 8 layers and 8 heads.}
    \label{tab:ol-depth}
    \end{table}
\section{Other Auxiliary Losses}

We experimented with other types of auxiliary losses on the object summary output from the Object Learner, 
always with the intention of improving the modality fusion by encouraging the object queries to attend to
both the modality streams.
However,  the results (\Cref{tab:auxloss}) show that they do not help achieve a better performance on downstream tasks, hence these other auxiliary losses are not included in the main paper. 
We list them below to illustrate approaches that don't benefit action or  hand state classification.
% Here we listed different types of losses we have explored and their performance on action recognition and hand contact state detection.
% According to our experiment results, they do not help achieve better results on downstream tasks than having single instance-level contrastive loss on object trajectories.
\subsection{Instance-level contrastive loss on visual transformation vectors}
% ** for each one give the intuition first on we are trying to capture/achieve before going into details **
To induce greater object-awareness, we train the object summary vectors to be able to pick out ‘correct’ visual dynamics from incorrect/synthetically generated ones. 
To this end, we introduce a contrastive loss with estimated `transformation vectors', which embeds the visual affinities of objects along the temporal dimension.
The `transformer vectors' computed from frames in a correct temporal order serve as positive samples, and the ones computed from temporally shuffled frames serve as negative samples in the loss.
The summary vectors are then tasked with associating each object to its ‘correct’ sample from the bag of positives and negatives. \\
% To ensure the summary vectors have captured the transformation of objects over time, 
% we introduce an estimated `affinity vector' from online feature maps and use it as the targets in the contrastive loss.
More specifically, given the visual feature maps of a clip and the object bounding boxes in it, 
we RoI-Pool the object features $w_j$ from each frame, where $j$ is the index of object. 
Based on these per-frame object features, we compute the `affinity vector' between frames by:

\begin{equation}
    \mathbf{\tilde{aff}}^i_j =
    \mathbf{w}^i_j \cdot \mathbf{\transpose{w}}^{i+1}_j ,
\end{equation}
\begin{equation}
    \mathbf{\tilde{aff}}^{shuffle,i}_j =
    \mathbf{w}^i_j \cdot \mathbf{\transpose{w}}^{k}_j , \quad k\neq i+1
\end{equation}
We embed the affinity vectors of an object along the temporal dimension into a transformation vector $z_j$. The encoding is done by using a small Transformer $g(.)$ with 2 layers and 4 heads.
\begin{equation}
    \mathbf{z_j} = g(\mathbf{\tilde{aff}}_j),
\end{equation}
\begin{equation}
    \mathbf{z}^{shuffle}_j = g(\mathbf{\tilde{aff}}^{shuffle}_j),
\end{equation}
\begin{equation}
\mathbf{\tilde{aff}}_j= (\mathbf{\tilde{aff}}^1_j, \mathbf{\tilde{aff}}^2_j, \hdots,\mathbf{\tilde{aff}}^{T'-1}_{j})
\end{equation}
\begin{equation}
    \mathbf{\tilde{aff}}^{shuffle}_j= Shuffle(\mathbf{\tilde{aff}}^1_j, \mathbf{\tilde{aff}}^2_j, \hdots,\mathbf{\tilde{aff}}^{T'-1}_{j})
\end{equation}
$z_j$ and $z^{shuffle}_j$ are used to compute the contrastive loss on object summary vectors $s_j$ as in:
\begin{equation}
    \mathcal{L}_{aff}=-\sum_{j}\Bigg[\log{\frac{ \exp(\transpose{s}_{j} \cdot {z}_{j} ) }
        {\sum_{k}{\exp(\transpose{s}_{j} \cdot {z}_{k})}+ \sum_{k}{\exp(\transpose{s}_{j} \cdot {z}^{shuffle}_{k})} } }\Bigg] 
    \label{eq:aux-aff}
\end{equation}

\subsection{Class-level contrastive loss on RoI-Pooled object vectors}
Based on the hypothesis that objects under the same action may have similar transformation of states, 
we design a contrastive loss to push these object summaries closer in the feature space. 
For each object $j$ and the action class label $l$ it is associated with. 
We apply a supervised contrastive loss on each object summary vector $s_j$,
where other vectors with the same class label $l$ serve as its positive samples, with different class labels are used as its negative samples.

\begin{equation}
    \mathcal{L}_{obj}=-\sum_{j}\Bigg[\log{\frac{\sum_{k}\exp(\transpose{s}_{j,l} \cdot {s}_{k,l} ) }
        {\sum_{k}\exp(\transpose{s}_{j,l} \cdot {s}_{k,l}) + \sum_{m,l'\neq l}{\exp(\transpose{s}_{j,l} \cdot {s}_{m,l'})} } }\Bigg] 
    \label{eq:aux-obj}
\end{equation}

\begin{table}[]
    \centering
    \resizebox{0.6\textwidth}{!}{%
    \begin{tabular}{c|c|c|l}
    \textbf{Loss}                   & \textbf{OL only} & \textbf{Backbone only} & \textbf{Final} \\ \hline
    $L_{traj}$                     &   71.0      &    72.0           &   73.6    \\
    $L_{traj} + L_{trans}$           &  71.0       &   72.1            &    73.5   \\
    $L_{traj} + L_{trans} + L_{obj}$ &  70.0       &    72.2           &    73.2   \\ \hline
    \end{tabular}%
    }
    \vspace{2mm}
    \caption{\textbf{Ablation on different types of auxiliary losses on SomethingElse.} Adding other auxiliary losses does not improve the action classification results. We choose to use a single contrastive loss on trajectories (Eq.1 in main paper) for simplicity.}
    \label{tab:auxloss}
    \end{table}
\section{Implementation Details}
\subsection{Data preprocessing}
During training, we sample clips of size $16\times 224\times 224$ uniformly from 
videos so that the temporal span of the clips cover the whole video.
Input images are normalized with mean and standard deviation 0.5, rescaling in
the range $[-1,1]$. For data augmentation, we apply random scale jittering from 
scale 180 to 256 such that the objects are not cropped out of the frames, random 
spatial cropping at size $224\times 224$, and random horizontal flips only to flipping-invariant 
classes (determined by class descriptions). We also use 
RandAugment~\cite{randaugment} with maximum magnitude 20 for color jittering. 
For inference, we use 3-crop evaluation following previous works~\cite{mformer,orvit}.

\subsection{Training}
We train the model with an AdamW~\cite{adamw} optimizer for 35 epochs with weight 
decay \num{1e-3}. The base learning rate is \num{3.75e-5}, decayed by 0.1 and 0.01 at epoch 20 and 30.
We use label smoothing~\cite{labelsmooth} with alpha 0.2 and mixed precision training.
The rate of DropConnect~\cite{dropconnect} in all attention layers is set to 0.2.
Due to limited compute resources (making joint end-to-end training infeasible), 
we first train the visual and trajectory backbone separately on corresponding training set, then freeze the visual backbone and fine-tune the trajectory backbone, Object Learner and Classification Module on 2 RTX 6000 GPUs with batch size 72.

\section{More Visualizations}
\begin{figure}[]
    \centering
    \includegraphics[width=0.95\linewidth]{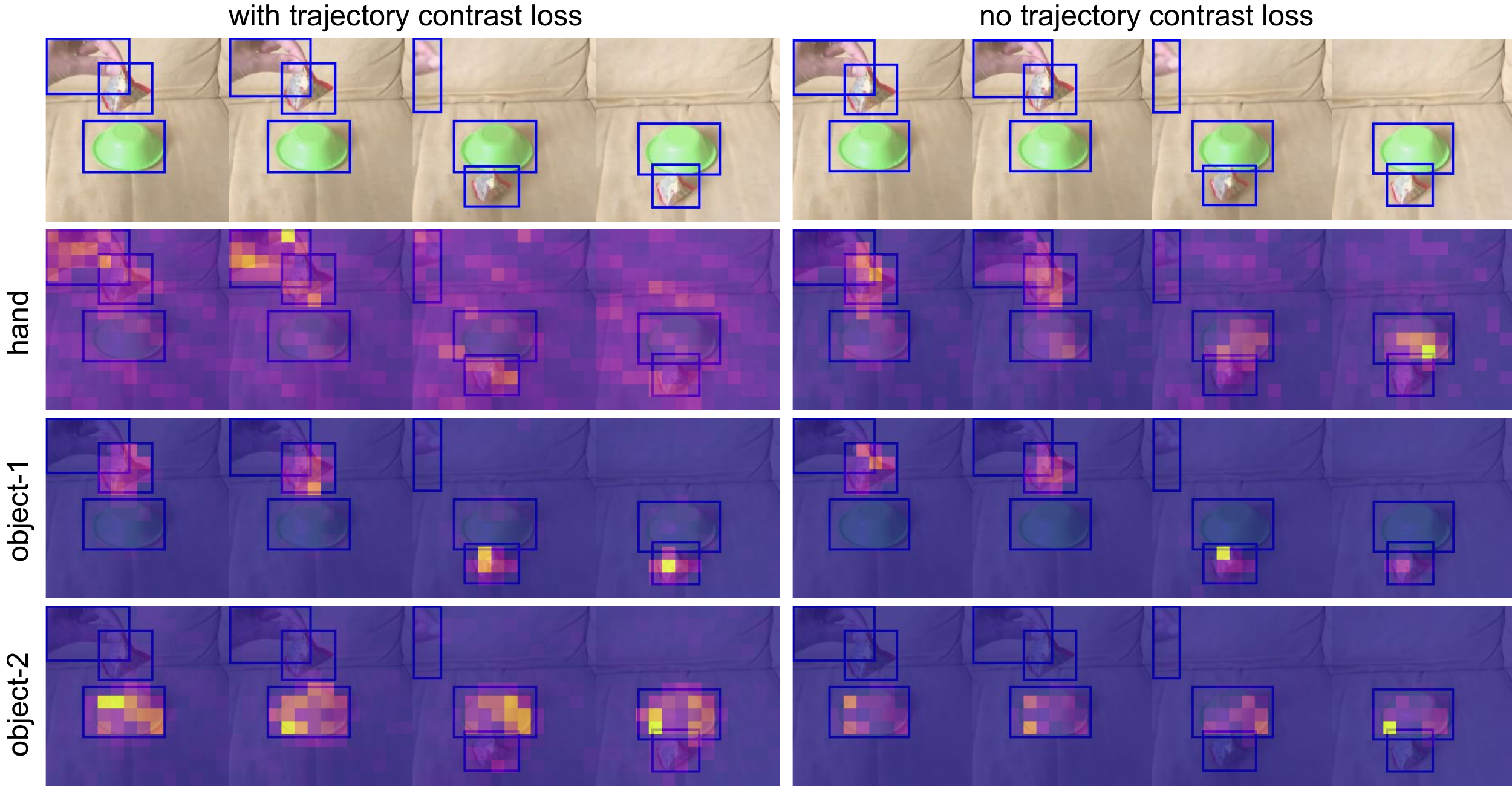}
    \includegraphics[width=0.95\linewidth]{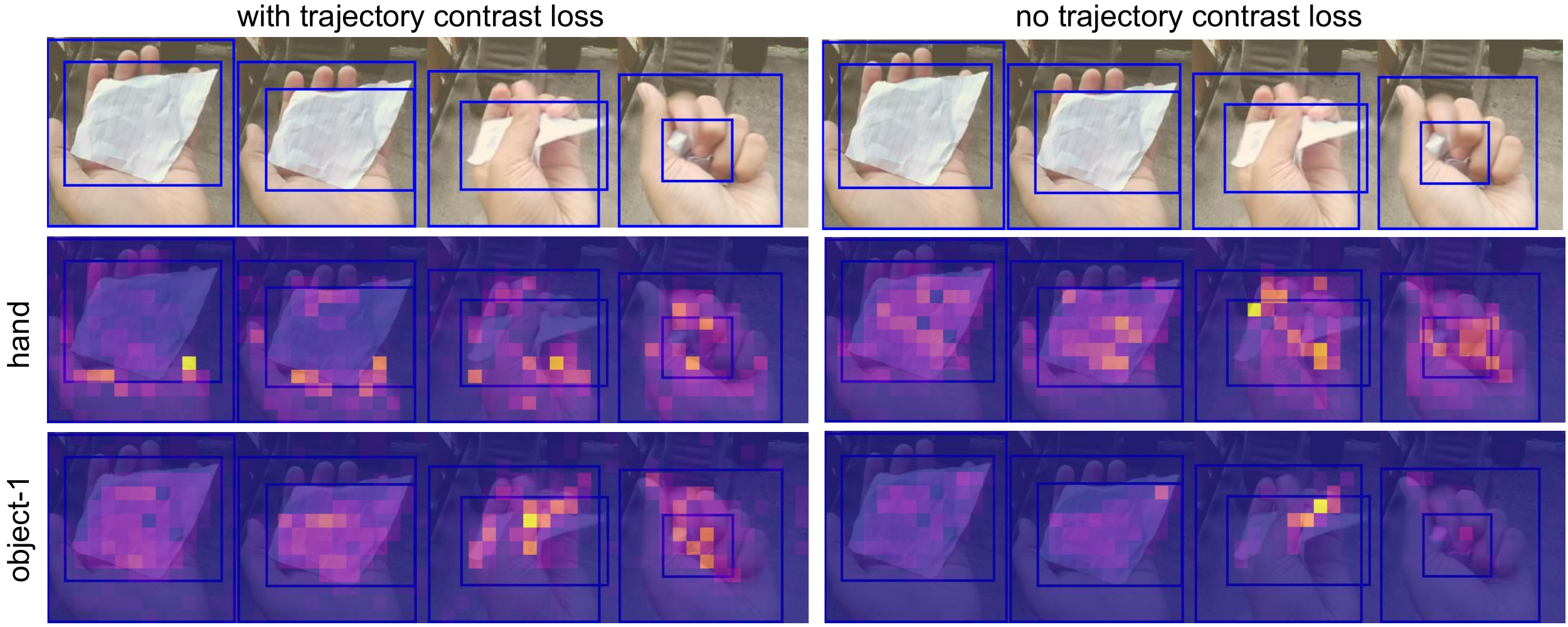}
    \includegraphics[width=0.95\linewidth]{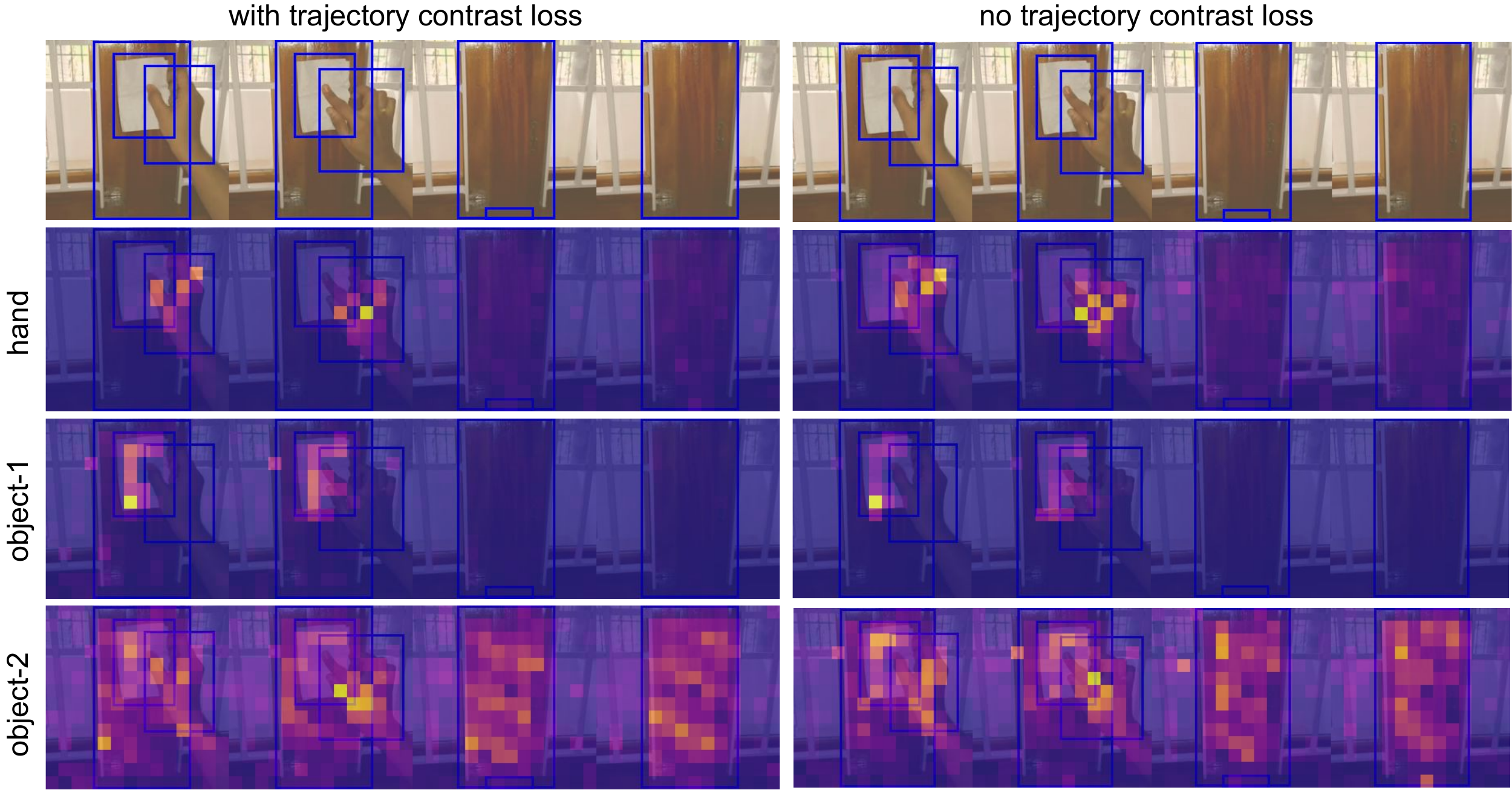}

    \vspace{-5mm}
    \caption{\textbf{Visualization of object-aware attention in Object Learner, from models trained with (Left) and without (Right) auxiliary loss.} 
    Attention of object queries on visual feature map is visualized above. 
    % \textbf{Left}: when the model is trained with auxiliary loss. 
    % \textbf{Right}: when the model is trained without auxiliary loss.
    Although in both cases the attention is object-centric, 
    the one trained without auxiliary loss does not always attend to the hands (middle figure),
    and has either weak or peaky attention on some parts of the objects (object-1, object-2 in the upper figure, object1-in the lower figure).
    While the one trained with the auxiliary loss always pays attention to the hand and even has  strong attention on the full objects.
    Brighter colors indicates higher attention scores.} 
    \vspace{-6mm}
    \label{fig:viz}
  \end{figure}

In the main paper we have shown some visualizations of object-aware attention in the Object Learner (Fig.4 in main paper), from models trained with and without auxiliary loss. 
The visualizations are done by plotting the attention scores from the last cross-attention layer. Here we add some more examples in \Cref{fig:viz}.

% \bibliographystyle{splncs}
% \bibliography{bib/shortstrings,bib/vgg_local,bib/vgg_other,bib/refs}

\end{document}